\documentclass[sigconf, nonacm]{acmart}

\definecolor{BrickRed}{RGB}{203,65,84}
\definecolor{RoyalBlue}{RGB}{65,105,225}

\usepackage{multirow}
\usepackage{fontawesome5} 
\usepackage{caption}
\usepackage{subcaption}
\usepackage{makecell}
\usepackage{colortbl}
\usepackage{graphicx} 
\usepackage{hhline}
\usepackage{enumitem}
\usepackage{titlesec}

\titlespacing*{\paragraph}{0pt}{0pt}{1em} % no vertical space before

\definecolor{gold}{RGB}{255, 215, 0} 
\definecolor{disco-llama}{RGB}{202, 120, 188}
\definecolor{llama}{RGB}{2, 115, 178}
\definecolor{leolm-mistral}{RGB}{222, 143, 2}

\AtBeginDocument{%
  }

\setcopyright{none}

\begin{document}

\title{Evaluating the Effectiveness of Direct Preference Optimization for Personalizing German Automatic Text Simplifications for Persons with Intellectual Disabilities}

% \author{\href{https://orcid.org/0009-0000-0876-621X}{Yingqiang Gao}}
% \authornote{Equal contribution.}
% \authornote{Corresponding author.}
% \affiliation{
%   \institution{University of Zurich}
%   \city{Zurich}
%   \country{Switzerland}}
% \email{yingqiang.gao@cl.uzh.ch}

\author{Yingqiang Gao}
\orcid{0009-0000-0876-621X}
\authornote{Equal contribution.}
\authornote{Corresponding author.}
\affiliation{
  \institution{University of Zurich}
  \city{Zurich}
  \country{Switzerland}
}
\email{yingqiang.gao@cl.uzh.ch}

\author{Kaede Johnson}
\orcid{0009-0002-6149-8167}
\authornotemark[1]
\affiliation{%
  \institution{EPFL}
  \city{Lausanne}
  \country{Switzerland}}
\email{kaede.johnson@epfl.ch}

\author{David Froehlich}
\orcid{0009-0007-8518-1566}
\affiliation{%
  \institution{capito.ai}
  \city{Graz}
  \country{Austria}}
\email{david.froehlich@capito.ai}

\author{Luisa Carrer}
\orcid{0000-0002-5784-8817}
\affiliation{%
  \institution{Zurich University of Applied Sciences}
  \city{Zurich}
  \country{Switzerland}}
\email{luisa.carrer@zhaw.ch}

\author{Sarah Ebling}
\orcid{0000-0001-6511-5085}
\affiliation{%
  \institution{University of Zurich}
  \city{Zurich}
  \country{Switzerland}}
\email{ebling@cl.uzh.ch}

\renewcommand{\shortauthors}{Gao and Johnson et al.}


% copy the authors back later


\begin{abstract}

Automatic text simplification (ATS) aims to enhance language accessibility for various target groups, particularly persons with intellectual disabilities.
Recent advancements in generative AI, especially large language models (LLMs), have substantially improved the quality of machine-generated text simplifications, thereby mitigating information barriers for the target group persons.
However, existing LLM-based ATS systems do not incorporate preference feedback on text simplifications during training, resulting in a lack of personalization tailored to the specific needs of target group representatives.

In this work, we extend the standard supervised fine-tuning (SFT) approach for adapting LLM-based ATS models by leveraging a computationally efficient LLM alignment technique—direct preference optimization (DPO).
Specifically, we post-trained LLM-based ATS models using human feedback collected from persons with intellectual disabilities, reflecting their preferences of paired text simplifications generated by mainstream LLMs.
Furthermore, we propose a  pipeline for developing personalized LLM-based ATS systems, encompassing data collection, model selection, SFT and DPO post-training,   and evaluation.
Our findings underscore the necessity of active participation of target group persons in designing personalized AI accessibility solutions aligned with human expectations.
This work represents a step towards personalizing inclusive AI systems at the target-group level, incorporating insights not only from text simplification experts but also from   target group persons themselves.\footnote{Code and data are available at \url{https://github.com/ZurichNLP/HF4ATS}.}

\end{abstract}




%\keywords{Do, Not, Us, This, Code, Put, the, Correct, Terms, for, Your, Paper}
\begin{teaserfigure}
  \includegraphics[width=\textwidth]{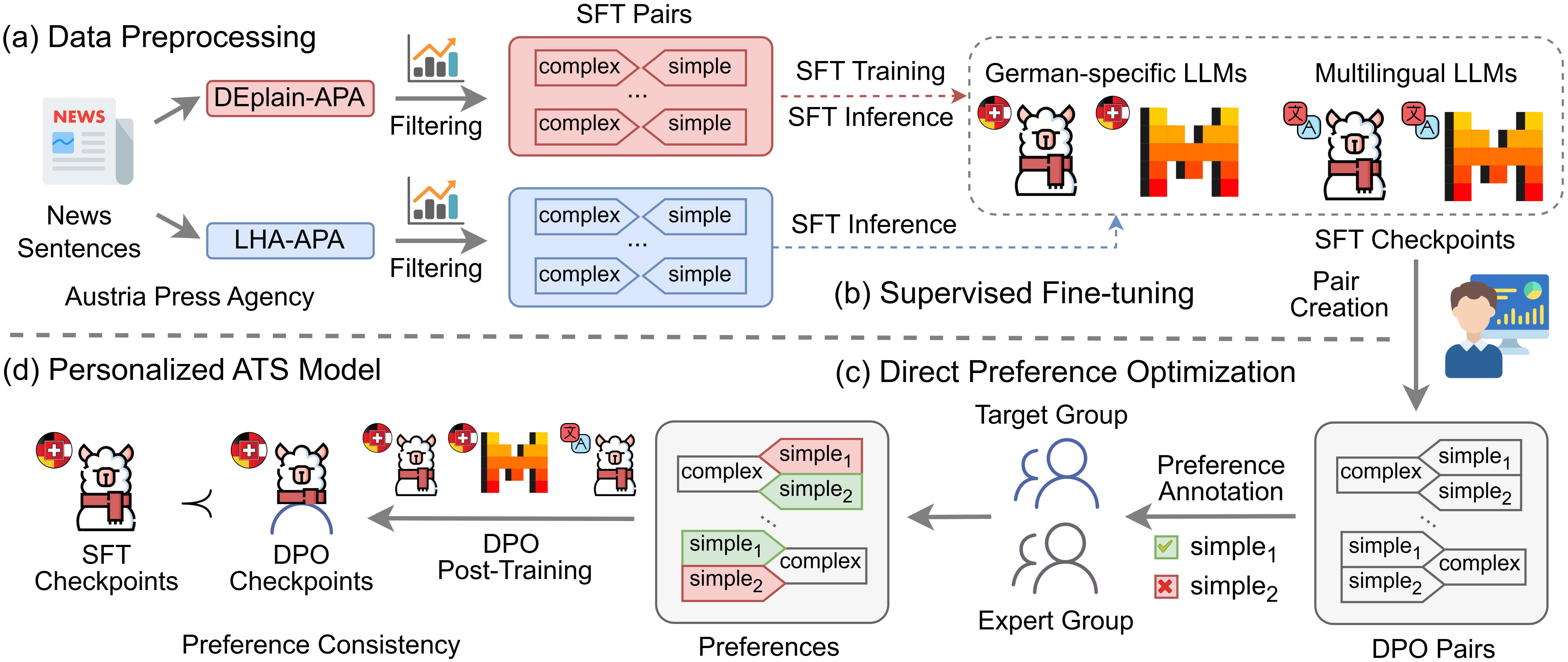}
  \caption{\textbf{Our personalization pipeline for LLM-based ATS models}. (a) \textbf{Data Filtering}: We selected high-quality sentence-level complex-simple pairs from two datasets; (b) \textbf{Supervised Fine-tuning}: We finetuned pre-trained German-specific and multilingual LLMs; (c) \textbf{Direct Preference Optimization}: We post-trained SFT checkpoints with human preference data collected from both target and expert group annotators; (d) \textbf{Evaluation}: We evaluated the DPO checkpoints against their SFT precursors for their alignment with human preferences.}
  \Description{This figure presents our four-stage pipeline for personalized automatic text simplification models, organized into panels (a) through (d). In panel (a) with the header text Data Preprocessing, sentence-level complex-simple pairs from two datasets are selected and subsequently filtered. These high-quality complex-simple pairs are passed on as input to panel (b), "Supervised Fine-tuning", where German-specific LLMs  and multilingual LLMs  are fine-tuned. These models produce new simplified outputs. A human evaluator icon leads to a box labeled "Pair Creation" that contains sentence pairs for comparison. In panel (c), "Direct Preference Optimization", the created sentence pairs are annotated by two user groups: a "Target Group" and an "Expert Group". The output is labeled as "DPO Pairs". In panel (d), "Personalized ATS Model", these pairs are used to train models further via DPO. The DPO checkpoints are compared with the SFT checkpoints.}
  \label{fig:teaser}
\end{teaserfigure}


\maketitle

\section{Introduction}

\subsection{Background}

Automatic text simplification (ATS) is a natural language processing (NLP) task that converts a standard-language text into an easier-to-understand version by improving text readability, increasing lexical and syntactic simplicity, and optimizing content complexity \citep{hansen2020intralingual, al2021automated}. 
Nowadays most often being tackled through AI approaches, ATS is oriented at diverse target groups, such as non-native language learners, persons with low literacy, and persons with intellectual disabilities. 
Among these target groups, persons with intellectual disabilities may encounter fundamental challenges in comprehending complex sentence structures, domain-specific jargon, implicit metaphors, and high information density, all of which can pose significant barriers to access and processing of daily-life information flows \citep{sauberli2024digital}.

Research in ATS has focused on enhancing the diversity of machine-generated simplifications—incorporating techniques such as text splitting, semantic paraphrasing, lexical substitution, and information deletion \citep{alva2020asset, maddela2021controllable, yamaguchi2023gauging}—as well as developing more robust evaluation metrics that better match with human judgment and perception \citep{maddela2023lens, cripwell2023simplicity, heineman2023dancing}.
With recent advancements in large language models (LLMs), ATS systems have become significantly more capable of generating high-quality text simplifications that are both fluent and adequate. The growing popularity of LLM-based ATS systems stems not only from their model supremacy in terms of  performance but also from their ease of integration and deployment as off-the-shelf solutions.

However, in practice, the perspectives of persons with intellectual disabilities are frequently overlooked in the development of inclusive AI technologies \citep{birhane2022power}. This is primarily because they are seldom consulted to provide feedback on AI-generated text simplifications based on their individual preferences. In addition, communication barriers associated with intellectual disabilities are often two-sided, meaning that not only do representatives of these target groups encounter difficulties in comprehending complex information but they may also face challenges in clearly expressing their  opinions \citep{cashin2024barriers}. As a result, if at all, their involvement in ATS research is typically limited to the final phase, namely human evaluation, where they provide feedback on system outputs that, in most cases, are not further refined based on their inputs.

Coming from current limitations in ATS research, we therefore identify two major challenges:

\paragraph{\textbf{Challenge 1}} Persons with intellectual disabilities are not constructively included in the implementation phase of AI-driven ATS research.

\paragraph{\textbf{Challenge 2}} Despite being a distinct target group, ATS models are rarely personalized for persons with intellectual disabilities, as their requirements differ from those of other target groups.

Existing ATS models are often trained on data curated by text simplification experts. As a result, they do not adequately address the aforementioned challenges related to the active participation of persons with intellectual disabilities.
Fortunately, various LLM alignment approaches, such as reinforcement learning from human feedback (RLHF; \citep{christiano2017deep}), have been introduced to enhance text generation by aligning LLM outputs with human values, expectations, and preferences \citep{ouyang2022training}. By integrating reinforcement learning into the post-training pipeline of LLMs, human feedback can serve as an active supervision signal, steering text generation toward a more human-centered and inclusive direction.
Nevertheless, traditional RLHF methods such as proximal policy optimization (PPO; \citep{schulman2017proximal})  are generally hard to train due to the need for (1) large-scale human preference data; (2) loading three LLMs (reference model, reward model, and policy model) simultaneously for training; and (3) sophisticated reward modeling and cumbersome hyperparameter tuning such as clipping threshold for policy updates. These limitations refrain PPO from many real-world applications such as in our case, personalizing LLM-based ATS models for the target group persons.

\subsection{Motivation}

In this work, we aim to personalize LLM-based ATS models for persons with intellectual disabilities (henceforth referred to as the target group) within a lightweight and cost-effective technical framework, incorporating human-in-the-loop (HITL; \citep{wu2022survey, mosqueira2023human}) participation.
As our primary personalization methodology, we propose investigating direct preference optimization (DPO; \citep{rafailov2023direct}), an LLM alignment algorithm that does not require explicit reward modeling. By integrating target group participants throughout all implementation phases and adhering to the validate-annotate-evaluate HITL principle, we aim not only to develop LLM-based ATS models that are post-trained on the preferences of the target group persons but also to establish an ethical and effective workflow for personalizing LLM-based ATS models.

The \textbf{main contributions} of our work are as follows:
(1) We developed a web application to collect human preference data from both target group participants and text simplification experts, ensuring minimal cognitive demand for interaction;
(2) We introduced HF4ATS, currently the largest human preference dataset containing annotated ATS pairs in German generated by mainstream LLMs;
(3) We open-sourced three LLM-based ATS models, post-trained with DPO on HF4ATS;
(4) We conducted extensive experiments to examine the impact of both model-level and data-level factors on the effectiveness of group-level personalization;
(5) We performed a detailed analysis of the outcomes of investigating preference consistency among the target group in the context of personalizing LLM-based ATS models.

The rest of this work is structured as follows: Section~\ref{sec:literature} reviews related literature on German ATS research and LLM personalization. Section~\ref{sec:data} details our data curation workflow, the model training process, and the personalization methodology. Section~\ref{sec:eval} describes both automatic and human evaluation protocols. Section~\ref{sec:results} presents the key research findings and provides an in-depth discussion of the experimental results. Finally, Section~\ref{sec:conclusion} summarizes the major takeaways of our study and highlights directions for future research.

\section{Related Works}
\label{sec:literature}
\subsection{German ATS Research}

As an NLP task that emerged in the late 1990s, ATS research initially relied on rule-based \citep{siddharthan2014hybrid, saggion2015making, suter2016rule} and statistical approaches \citep{bach2011tris, xu2016optimizing, qiang2019unsupervised}. Rule-based methods employed look-up tables for lexical and syntactic operations, while statistical approaches framed text simplification as a sequence-to-sequence task, often modeled using statistical machine translation.
Recent advancements in LLMs have significantly transformed state-of-the-art ATS systems, enabling an end-to-end approach without any feature engineering. This shift has progressed from encoder-decoder architectures to decoder-only models, driven by the enhanced computational efficiency and scalability of modern LLMs. As a result, decoder-only LLMs have become general-purpose problem solvers, redefining the learning paradigm for ATS.

While most ATS models were trained on English data \citep{scarton2018learning, sheang2021controllable, agrawal2024text}, German ATS research has gained increasing attention in recent years, driven by active political and legal initiatives in German-speaking countries \citep{ebling2022automatic}. 
Notable examples include the \textit{Accessible Information Technology Regulation} (\textit{Barrierefreie-Informationstechnik-Verordnung}, BITV 2.0, 2011) in Germany, national
 action plans on disability in Austria \citep{bmsgpk2022disability},
and the ratification of the \textit{United Nations Convention on the Rights of Persons with Disabilities} (UN-CRPD) in Germany (2009), Austria (2008), and Switzerland (2014).
%Notable examples include the \textit{Accessible Information Technology Regulation} (\textit{Barrierefreie-Informationstechnik-Verordnung}, BITV 2.0, 2011) in Germany and the ratification of the \textit{United Nations Convention on the Rights of Persons with Disabilities} (UN-CRPD) in Germany (2009), Austria (2008), and Switzerland (2014). 
These efforts have significantly advanced German ATS research, particularly in areas such as dataset construction \citep{klaper2013building, battisti2020corpus, sauberli2020benchmarking, gonzales2021new, aumiller2022klexikon, seiffe2022subjective, toborek2023new, stodden2023deplain, kloser2024german}, alignment of  texts \citep{spring2022ensembling, spring2023analyzing}, and training of models \citep{spring2021exploring, anschutz2023language, hewett2024elaborative}.

%\textit{Barrier-Free Information and Communication} (\textit{Barrierefreie Information und Kommunikation}; BIK) in Austria, %

Among existing German ATS models, BART \citep{lewis2020bart}, T5 \citep{raffel2020exploring} and their multilingual variants \citep{liu2020multilingual, xue2021mt5} are the most commonly used base models, while benchmarking against commercial LLMs like GPT-3.5 or GPT-4 has also become a standard practice.
In this work, we focus on fine-tuning and post-training mainstream open-source LLMs, such as Llama \citep{touvron2023llama, dubey2024llama} and Mistral \citep{jiang2023mistral}, to evaluate their ability to generate human-preferred text simplifications.

\subsection{LLM Personalization}
\label{sec:LLM-personalization}
As summarized in the literature \citep{parthasarathy2024ultimate}, there are three stages in the traditional route toward superlative text generation:

\paragraph{\textbf{Large-scale Pre-training}.} In this stage, auto-regressive LLMs (also known as causal LLMs) acquire world knowledge from large-scale corpora through self-supervised next-token prediction. This process optimizes a cross-entropy objective to maximize the accumulated log-likelihood, enabling the pre-trained language models (PLMs) to develop a deeper linguistic and contextual understanding of language.

\paragraph{\textbf{Supervised Fine-tuning}.} PLMs undergo further fine-tuning on task-specific datasets to improve their capability in task comprehension and instruction adherence. The resulting supervised fine-tuned (SFT) models are optimized for both task execution and instruction following, enhancing their overall effectiveness in downstream NLP applications.

\paragraph{\textbf{Preference Learning}.} After acquiring world knowledge and developing the ability to follow human instructions, the final stage involves post-training SFT models on human preference data to match their responses with human expectations. At this stage, SFT models are personalized to enhance their alignment with human preferences, ensuring they generate ethical, helpful, and informative responses with greater robustness.

Preference learning is a key approach to personalizing LLMs, alongside methods such as personalized prompting and personalized adaptation \citep{liu2025survey}. It is the only method that directly incorporates subjective human feedback into the learning process \citep{zhao2025do}. While on one hand personalized prompting and personalized adaptation require explicit user profile data to guide text generation, which might cause ethical issues, on the other hand preference learning relies solely on human feedback data, which is much easier to anonymize.

Reinforcement learning from human feedback (RLHF) methods, such as proximal policy optimization (PPO \citep{schulman2017proximal}), utilize preference data to train an explicit reward model that guides the alignment process. As an on-policy RLHF approach which requires on-the-fly training data generation, PPO has demonstrated remarkable performance in conversational and coding tasks. However, it requires loading three LLMs simultaneously—a reference model $\pi_{\mathrm{ref}}$, a policy model $\pi_\theta$, and a reward model $R_{\psi}(\cdot)$—making it particularly computation-intensive and hard to train. 
Other related studies \citep{zhang2017sentence, nakamachi2020text} have explored the use of reinforcement learning within lightweight encoder-decoder generation pipelines for ATS. However, there remains a need for an explicit reward function that effectively integrates supervised training signals which contains no human preferences. 

On the other hand, methods such as direct preference optimization (DPO; \citep{rafailov2023direct}) eliminate the need for explicit reward modeling by learning directly from preference data, making them much more lightweight and computationally efficient \citep{ivison2024unpacking}. Moreover, as an offline method, DPO enables the pre-curation of preference data, thereby streamlining the data collection process while ensuring direct involvement with the target group persons.

In the context of ATS, given a preference dataset $\mathcal{D}$ consisting of triples of $(x, y_w, y_l)$, where $x$ is the complex text, $y_w$ is the LLM-generated text simplification preferred by humans, and $y_l$ is the LLM-generated text simplification dispreferred by humans, DPO aims at learning a policy model $\pi_\theta$ that assigns a higher preference score to the preferred text simplification $y_w$. The human preference can be therefore modeled as probabilistic ranking with the Bradley-Terry model \citep{bradley1952rank}:  
\begin{align*}
    P(y_w \succ y_l | x) & = \sigma (R_\psi (x, y_w) - R_\psi(x, y_l)) \\
    & = \frac{\exp (R_\psi(x,y_w))}{\exp(R_\psi(x,y_w)) + \exp(R_\psi (x,y_l))},
\end{align*}
where $\sigma(\cdot)$ is the sigmoid function.
With some reparametrization trick that essentially gets rid of the explicit reward modeling $R_\psi(x,y)$ and estimates the implicit reward $\hat{r}(x,y)$ directly from the training samples, the DPO training objective becomes
\begin{align*}
    & \mathcal{L}_{\mathrm{DPO}}(\pi_\theta;\pi_{\mathrm{ref}}) = \\
    & - \mathbb{E}_{(x,y_w,y_l) \sim \mathcal{D}} \left[ \log \sigma \left( \beta \log \frac{\pi_\theta (y_w | x)}{\pi_{\mathrm{ref}}(y_w | x)} - \beta \log \frac{\pi_\theta (y_l | x)}{\pi_{\mathrm{ref}}(y_l | x)} \right) \right],
\end{align*}
where the parameter $\beta$ actively regulates the deviation of the policy model $\pi_\theta$ from the reference model $\pi_{\mathrm{ref}}$, ensuring that the log-odd differences remain within a controlled range. This log-odd difference between the preferred and dispreferred text simplification is the so-called implicit \textbf{reward margin}
\begin{align*}
\label{eq:implicit_reward}
    \hat{r}(x,y_w, y_l) = \beta \left(  \log \frac{\pi_\theta (y_w | x)}{\pi_{\mathrm{ref}}(y_w | x)} - \log \frac{\pi_\theta (y_l | x)}{\pi_{\mathrm{ref}}(y_l | x)} \right),
\end{align*}
which can be directly estimated from the training instances.
A common practice when post-training with DPO is to initialize both $\pi_{\mathrm{ref}}$ and $\pi_\theta$ with the SFT model checkpoint and freeze $\pi_{\mathrm{ref}}$ during post-training, so that the gradient $\nabla_\theta \mathcal{L}_{\mathrm{DPO}}(\pi_\theta;\pi_{\mathrm{ref}})$ will only be back-propagated to the policy model $\pi_\theta$ for updating its parameters.

As demonstrated by \citep{ivison2024unpacking}, high-quality preference data is the primary factor driving performance improvements in DPO post-training. While methods such as personalized prompting and personalized adaptation enable LLMs to generate user-specific responses based on personal attributes—such as gender, social relationships, occupation, and personal interests \citep{liu2025survey}—we argue that these approaches remain largely infeasible for personalizing LLMs for persons of the target group of interest here. This limitation is primarily due to ethical and legal constraints, as the construction of user profiles for persons with intellectual disabilities is considered unethical and, in many jurisdictions, legally prohibited.

In this work, we focus on user-agnostic, group-level LLM personalization with preference learning, aiming to develop LLM-based ATS systems that cater to the needs of target group persons as a whole. Our approach relies solely on preference data over LLM-generated text simplifications collected from target group persons, ensuring group-level personalization without any user profiling.

Along with our core methodology DPO, we proposed the following \textbf{Research Questions} (RQs) to be investigated in this work:

\paragraph{\textbf{RQ1}} Can DPO post-training with pairwise human preferences further improve the quality of ATS, as measured by automatic evaluation metrics?

\paragraph{\textbf{RQ2}} To what extent do factors such as preference source, information equality, and generalization of LLMs influence the effectiveness of DPO post-training? 

\paragraph{\textbf{RQ3}} Despite these challenges, can DPO post-training ultimately enable successful group-level personalization of ATS models? 

Next, in Section~\ref{sec:data}, we introduce our research pipeline, including 1) the introduction of HF4ATS, our curated human preference dataset designed for post-training German LLM-based ATS models, 2) the LLM models we used for SFT and DPO, and 3) the training and hyper-parameter tuning of the LLM-based ATS models.

\section{Data, Model, and Methods}
\label{sec:data}

We introduce Human Feedback for Automatic Text Simplification (HF4ATS), a dataset designed to enhance German ATS through learning with human preferences. 
To the best of our knowledge, HF4ATS is the first and largest German-language preference dataset collected directly from the target group for this purpose.
HF4ATS consists of two key datasets: (1) HF4ATS-SFT ($\mathcal{D}_{\text{SFT}}$), a dataset of complex-simple sentence pairs suitable for fine-tuning German LLM-based ATS models, and (2) HF4ATS-DPO ($\mathcal{D}_{\text{DPO}}$), an ATS preference pair dataset annotated by native German speakers. $\mathcal{D}_{\text{DPO}}$ can be adapted for use in several preference alignment frameworks. In this work, we use it to post-train ATS models with DPO.

\subsection{SFT Phase}

\subsubsection{SFT Model Selection}
To develop robust ATS models, we started with four state-of-the-art LLMs as backbones, prioritizing models that (1) were either multilingual or specifically tuned for the German language, ensuring sufficient German linguistic competence, and (2) had been instruction-tuned to effectively follow simplification guidelines. Based on these criteria, we chose LLMs with approximately 8 billion parameters, including Llama-3.1-8B-Instruct, DiscoLeo-Llama-3-8B-Instruct, Mistral-7B-Instruct, and LeoLM-Mistral-7B-Chat. Among these, DiscoLeo-Llama and LeoLM-Mistral have been explicitly pre-trained on German texts, further enhancing their suitability for ATS in German.

\subsubsection{Data Filtering}

We curated HF4ATS-SFT from \textsc{DEplain} \citep{stodden2023deplain}, a parallel (standard-language/simplified-language) simplification dataset containing professionally written and manually aligned  simplifications. \textsc{DEplain} consists of two sub-collections: \textsc{DEplain-APA} and \textsc{DEplain-Web}. While the latter is derived from web-crawled documents consisting of non-news texts, the former comprises text from Austrian Press Agency (APA) news items published from May 2019 to April 2021 and covering a diverse range of topics. These topics include politics, crime, weather, economics, zoo births, and the coronavirus pandemic. 
Overall, \textsc{DEplain-APA} contains 13,122 manually aligned sentence pairs from 483 documents pairs classified as A2 or B1 under the Common European Framework of Reference for Languages (CEFR). Given its diverse topic coverage, we selected \textsc{DEplain-APA} as the base for HF4ATS-SFT. When discussing \textsc{DEplain}-APA sentences, we refer to text from B1 articles as ``complex'' text and text from A2 articles as ``simple'' or ``simplified'' text.

To ensure the inclusion of high-quality pairs during SFT, we incorporated the following data filtering steps: First, we excluded pairs with many-to-many or many-to-one mappings (850 pairs, accounting for 6.4\% of the total pairs in \textsc{DEplain-APA}), retaining only those with one-to-one or one-to-many mappings. This selection ensured a focus on pairs that did not introduce overly dense information. Second, we sought to remove pairs in which the simplified text was not entailed by the corresponding complex text. This lack of entailment arises in \textsc{DEplain}-APA due to insufficient sentence-level context. One example is as follows:
    
\textit{Complex (from a B1 news article)}: \textit{Es gibt aber grosse Unterschiede}. (English: \textit{But there are big differences}.)
    
\textit{Simple (from an A2 news article)}: \textit{Nicht in jedem Vanille-Eis ist gleich viel Luft drin}. (English: \textit{Not every vanilla ice cream contains the same amount of air}.)

Without sufficient context, the simplification above cannot be reliably inferred from the original text. This is because the complex-simple pairs, despite being manually aligned, originate from entire articles rather than an isolated sentence-to-sentence(s) simplification task. Simplified text in \textsc{DEplain}-APA may therefore rely on contextual cues from a source article that are not present in the paired complex text. If an excessive number of such instances were included during SFT, our models could have developed a tendency to generate hallucinated, non-faithful, or even unethical content. 

To implement this filter, we employed a semantics-based approach by computing the cosine similarity between complex and simplified texts. Specifically, we utilized a Sentence-BERT model \citep{reimers2019sentence} fine-tuned for the German language\footnote{Available at \url{https://huggingface.co/T-Systems-onsite/cross-en-de-roberta-sentence-transformer}, MIT license.}. Based on empirical analysis, we filtered out 591 text pairs with cosine similarity scores below the threshold of 0.5, which is a deliberately relaxed filter given the involvement of human input when creating preference pairs at a later stage.

Third, we removed pairs in which the simplified texts were overly similar to the complex texts, as indicated by a high degree of N-gram overlap. The following pair is an extreme example:

\textbf{Complex}: \textit{Integration bedeutet, also dass jemand dazugeh\"ort}. (English: \textit{Integration means that someone belongs}.)

\textbf{Simple}: \textit{Integration bedeutet also, dass jemand dazugeh\"ort}. (English: \textit{Integration means that someone belongs}.)

Including such pairs in the SFT process could have reduced the diversity of simplifications generated by LLMs. This reduction in diversity may have, in turn, triggered a cascade effect during DPO post-training, as training on preferences derived from overly similar simplifications could lead to insufficiently informative reward discrepancies. To address this issue, we applied an additional heuristic filter, removing 2,322 pairs whose F1 score across ROUGE-1, ROUGE-2, and ROUGE-L \citep{lin2004rouge} exceeded a threshold of 0.8.
%Figure~\ref{fig:deplain_rouge} illustrates the N-gram overlap in the original \textsc{DEPlain-APA} pairs, demonstrating that high overlap is prevalent despite manual alignment.   

%\input{figures/deplain-rouge}

Lastly, to ensure task completion would only introduce a moderate amount of information, we removed 116 pairs in which the simplification exceeded 30 words.

The remaining set amounted to 9,359 pairs following this four-step data filtering process. To create the training, development, and test sets for SFT, we applied a stratified approach with a 70\%-15\%-15\% split. Specifically, we randomly allocated 3,600 train, 800 development, and 800 test pairs from a subset of 5,200 pairs purposefully sampled to balance the sentence length distribution. To achieve this, we employed Gaussian sampling based on the word count of the complex texts, aiming for an average length of 13 words (up from 11.24 words for the 9,359 pre-sample \textsc{DEplain-APA} pairs).

Formally, for a given complex text $x$, the sampling weight $w_x$ is defined as
\begin{align*} 
w_{x} = \exp{ \left(- \frac{(|x| - 15)^2}{2 \cdot \sigma^2} \right)}, 
\end{align*}
where $|x|$ denotes the word count of the complex text and the standard deviation $\sigma$ is set to 3. This sampling formula yielded a less skewed word count distribution for complex texts in our final subset of 5,200 pairs.

%Figure~\ref{fig:deplain_presft_wcs} shows that after the Gaussian sampling [JUSTIFICATION NEEDED]

%\input{figures/deplain-wcs}

\subsubsection{Input Prompts}

In collaboration with a German-speaking text simplification expert, we developed ten prompts for  SFT. Eight of these prompts were later re-used for DPO post-training. The prompts address different aspects of our specific ATS use case with variety in their approach. The components that appeared in one or more prompts were as follows:

\begin{itemize}[left=0pt]
    \item Description of the target audience (German-speaking persons with intellectual disabilities)
    \item Goal of easy language (German: \textit{Leichte Sprache})
    \item Suggestion of text simplification operations (including adding, removing, reordering, replacing, and splitting) according to the recommendations for German Easy Language (German: \textit{Empfehlungen f\"ur Deutsche Leichte Sprache}\footnote{\url{https://www.din.de/de/mitwirken/normenausschuesse/naerg/e-din-spec-33429-2023-04-empfehlungen-fuer-deutsche-leichte-sprache--901210}}, and as in \citet{maass2015leichte})
    \item One-shot prompting with one concrete example
    \item Two-shot prompting with two concrete examples
\end{itemize}

All input prompts included at least one of the aforementioned components. The inclusion of few-shot prompts leveraged in-context learning benefits for SFT \citep{chen2022meta, mosbach2023few}.

Appendix~\ref{sec:prompts} provides a comprehensive list of the prompt templates used for SFT (and DPO). To ensure a consistent response format and facilitate post-processing, all prompts include the instruction: ``\textit{Bitte gib nur die Vereinfachung an, ohne Einleitung, Alternativen und Kommentare}''. (English: ``\textit{Please provide only the simplification, without introduction, alternatives, or comments}''.)

\subsubsection{SFT Training} 
\label{subsubsec: sft_training}

We implemented SFT using the 3,600 training pairs from HF4ATS-SFT. Following the findings of \citep{zhou2023lima}, which suggest that the optimal SFT checkpoint may emerge after a few thousand training instances, we periodically evaluated model performance on the 800-pair development set. Specifically, evaluations were conducted after every 400 training instances during cross-model comparisons and every 448 training instances during hyperparameter tuning.

To pad the input texts for LLMs, we set the padding token to <|\texttt{finetune\_right\_pad\_id}|> for Llama-3.1-8B-Instruct, <\texttt{unk}> for the two Mistral models, and left it unchanged for DiscoLeo-Llama-3-8B-Instruct. Input padding was consistently applied to the right side of the prompts.

Research has shown that full-prompt tuning—tuning where instruction tokens are included in the training loss calculation—can enhance performance for open-ended tasks when the average ratio of prompt token count to completion token count exceeds five and the number of training instances amounts to a few thousand \citep{shi2024instruction}. These two conditions are met by HF4ATS-SFT. Therefore, to increase robustness, we adopted a mixed strategy and trained separate models with full-prompt tuning and completion-only tuning.

\subsubsection{SFT Checkpoint Evaluation}
We employed an offline evaluation strategy to assess 36 SFT checkpoints saved at regular training intervals. The evaluation focused on performance across three key dimensions:
\begin{itemize}[left=0pt]
    \item \textbf{Simplification Quality}: We evaluated using BERTScore \citep{zhang2019bertscore} (F1 measure), BLEU \citep{papineni2002bleu}, and SARI \citep{xu2016optimizing} on the development set. SARI was selected as the most salient metric for evaluation. 
    \item \textbf{Simplification Readability}: We assessed average word count, Flesh Reading Ease \citep{kincaid1975derivation}, and Wiener Sachtextformel Variant 4 ($\text{WSTF}_4$; \citep{bamberger1984lesen})  (English: Vienna Formula) on the development set. $\text{WSTF}_4$ is a reference-free readability metric for German texts defined as follows:
    \begin{align*}
        \text{WSTF}_4 = 0.2744 \times MS + 0.2656 \times SL - 1.693, 
    \end{align*}
    where $MS$ represents the percentage of words containing more than three syllables and $SL$ denotes the average word count per sentence. A $\text{WSTF}_4$ score of 4 indicates a very simple text, while a score of 15 indicates a very complex text. We selected  $\text{WSTF}_4$ as the most salient readability metric over Flesch Reading Ease because it was specifically designed for non-fiction, German-language text. Both Flesch Reading Ease and $\text{WSTF}_4$ were computed using the Python package \texttt{Textstat}\footnote{Available at \url{https://github.com/textstat/textstat}, MIT license.}, configured for German language settings.
    \item \textbf{SFT Implementation Quality}: We assessed the mirror rate—the proportion of generated simplifications identical to the complex input (computed on lowercase, alphabetic characters only)—and cross-entropy validation loss on our development set. These metrics were intended to assess our implementation's ability to generalize by monitoring effective task completion and potential overfitting.
\end{itemize}

% \input{figures/grid_search}

%Figure~\ref{fig:sft_paramix_grid_search} presents the greedy grid search results for SFT hyperparameters. 
Based on SARI and $\text{WSTF}_4$ performance, we selected the hyperparameter configuration with a gradient accumulation step size of 1 and a learning rate of 1e-4 for cross-model comparison. We then trained all four models with this configuration, implementing full-prompt tuning generally and completion-only tuning for the two Llama-based models. Figure~\ref{fig:sft_modelmix_grid_search} in Appendix~\ref{sec:training-details} illustrates the cross-model comparison.

We sought diversity in model and training regime for our DPO reference checkpoints. Based primarily on SARI, $\text{WSTF}_4$, average simplification length, and mirror rates, we selected the following three SFT checkpoints: 
\begin{enumerate}[left=0pt]
    \item DiscoLeo-Llama-3-8B-Instruct after 2,800 training steps of full-prompt tuning;
    \item Llama-3.1-8B-Instruct after 2,400 training steps of completion-only tuning;
    \item LeoLM-Mistral-7B-Chat after 1,600 training steps of full-prompt tuning.
\end{enumerate}

\subsection{DPO Phase}

\subsubsection{DPO Preference Pairs Creation} 

After training the SFT models on HF4ATS-SFT, we created the ATS pairs for HF4ATS-DPO by first performing inferences with the selected SFT checkpoints. These simplifications were generated  for 8,000 complex texts curated from two sources: (1) 3,200 complex texts sampled from all \textsc{DEplain} pairs not included in HF4ATS-SFT, denoted $\mathcal{D}_{\text{\textsc{DEplain}}} \backslash \mathcal{D}_{\text{SFT}}$, and (2) 4,800 complex texts sampled from the APA-LHA dataset \citep{spring2021exploring}, denoted $\mathcal{D}_{\text{LHA}}$. APA-LHA comprises automatically aligned sentence-level complex-simple text pairs pulled from APA news items classified as A2 and above. While we excluded this dataset from SFT because its automatic alignments could have induced hallucinations, we did involve its complex sentences during pair creation because its topic distribution is similar to that of \textsc{DEplain} (in fact, the two datasets shared some complex sentences).

We applied Gaussian sampling with different weighting schemes to the two HF4ATS-DPO inference sources. From the \textsc{DEplain} subset we sampled a complex text $x$ with a Gaussian weight $w_x$ defined as 
\begin{align*}
    w_{x \sim \mathcal{D}_{\text{\textsc{DEplain}}} \backslash \mathcal{D}_{\text{SFT}}} = \exp \left( -
 \frac{\left(|x| - \mu_{\mathcal{D}_{\text{\textsc{DEplain}}} \backslash \mathcal{D}_{\text{SFT}}} \right)^2 }{2 \cdot \sigma^2}\right),
\end{align*}
where the mean 
\begin{align*}
    \mu_{\mathcal{D}_{\text{\textsc{DEplain}}} \backslash \mathcal{D}_{\text{SFT}}} = \frac{\sum_{x' \in \mathcal{D}_{\text{\textsc{DEplain}}} \backslash \mathcal{D}_{\text{SFT}}}|x'|}{|\mathcal{D}_{\text{\textsc{DEplain}}} \backslash \mathcal{D}_{\text{SFT}}|}
\end{align*}
corresponds to the average word count of complex texts from the leftover \textsc{DEplain} subset, with $|\mathcal{D}_{\text{\textsc{DEplain}}} \backslash \mathcal{D}_{\text{SFT}}|$ denoting the subset's size and $|x|$ denoting a given complex text's word count. 
From APA-LHA, $x$ was sampled with weight $w_x$ defined as 

\begin{minipage}{\columnwidth}
\centering
\resizebox{0.95\columnwidth}{!}{$
\begin{aligned}
    w_{x \sim \mathcal{D}_{\text{LHA}}} = \exp \left(- \frac{\left( |x| - \left( \mu_{\mathcal{D}_{\text{LHA}}}+ \eta \cdot (\mu_{\mathcal{D}_{\text{LHA}}} - \mu_{\mathcal{D}_{\text{\textsc{DEplain}}} \backslash \mathcal{D}_{\text{SFT}}}) \right) \right)^2}{2 \cdot \sigma^2} \right),
\end{aligned}
$}
\end{minipage}
where the mean 
\begin{align*}
    \mu_{\mathcal{D}_{\text{LHA}}} = \frac{\sum_{x' \in \mathcal{D}_{\text{LHA}}}|x'|}{|\mathcal{D}_{\text{LHA}}|}
\end{align*}
represents the average word count of complex texts from APA-LHA and $\eta = 4,800/8,000$ is a scaling factor reflecting the share of LHA-APA (as opposed to leftover \textsc{DEplain}) complex sentences present in the 8,000 instance inference set.

We generated 20 text simplifications per SFT checkpoint for all 8,000 complex sentences. One of eight prompts was assigned to each complex sentence at random (see Appendix~\ref{sec:prompts}), and we varied temperature and the top-p sampling parameter to achieve inference variety with four decoding configurations. 

Once inference was completed, 13 proficient German-speaking human pair creators (CEFR-level C1 and above) with strong backgrounds in computational linguistics research reviewed the automatic simplifications to construct ATS pairs for preference annotation.

Pair creators were trained with the following rubric to ensure high-quality ATS pairs:

\begin{itemize}[left=0pt]
\item \textbf{Entailment}: Pair creators verified that the complex sentence entailed the simplification. Because adding information is a valid simplification strategy, pair creators were allowed to make an exception in an unambiguous situation (e.g., a simplification that identifies the ``Democratic candidate for the 2020 U.S. Presidential election'' as ``Joe Biden'').
\item \textbf{Equal information}: Pair creators prioritized text simplifications that conveyed the same amount of information. Creators indicated via the creation tool whether each pair met this condition.
\item \textbf{High simplification quality}: Pair creators prioritized simplifications that adhered to German language rules and were accessible for persons with intellectual disabilities. We also actively asked pair creators to avoid simplifications that were potentially non-ethical or non-faithful. 
\item \textbf{High simplification diversity}: Pair creators prioritized selecting two simplifications that differed in their applied simplification strategies (e.g., deletion, paraphrasing, or sentence splitting).
\end{itemize}

To facilitate the pair creation process, we developed an intuitive Python script that enabled human pair creators to review 20 inferences for a complex sentence, select two to pair together, and indicate whether their selected simplifications had equal levels of information. Pairs could only be created with two inferences from the same winning SFT checkpoint, and pair creators were able to skip inference sets if no suitable pairs could be identified. Along with the order of complex sentences, the order in which the three winning SFT checkpoint's inference sets appeared was randomized. Each complex sentence was only shown until one appropriate pair was created or all SFT checkpoint inference sets were skipped by the pair creator. Additionally, the SFT checkpoint responsible for each inference set remained masked during annotation.

 We gathered 3,037 ATS pairs for preference annotation, each with a unique complex sentence from the 8,000 complex sentences we sampled from APA-LHA and leftover \textsc{DEplain} data.

\subsubsection{DPO Preference Pair Annotation} 

Many crowd-sourced data annotation tools involve complex human-computer interaction (HCI) designs that can be cognitively overwhelming for participants from the target group. To reduce cognitive load, we developed a minimal, lightweight, and user-friendly web application to collect preferences from target group participants. 
Given  evidence from previous research \citep{sauberli2024digital} indicating that target group participants often experience reading difficulties, 
%we further minimized cognitive demands by omitting each pair of automatically simplified sentences', i.e., each preference pair's, corresponding original text in the user interface of the web tool for all target group participants. 
we further minimized cognitive demands for all participants in the target group, by omitting the original standard German text for each pair of automatically simplified sentences, i.e., for each pair of preferences, from the user interface of the web tool.
Instead, target group participants were simply asked to indicate their preference between two presented automatic simplifications, based solely on which version they found easier to understand. 
%We show this design in Figure~\ref{fig:web-tool-target}.

% \input{figures/dpo-anno-tool-target}

The preferred text simplification was highlighted with a light green background once the participant selected the corresponding button, ``Diesen Text verstehe ich besser'' (English: ``I understand this text better''). After completing the current pair, participants could freely navigate to the previous or next pair using the ``Zur\"uck'' (English: ``Back'') and ``Weiter'' (English: ``Next'') buttons. Annotations could be submitted at any time by clicking the ``Abschicken'' (English: ``Send'') button.

To ensure compliance with data privacy and protection regulations, both the front end (user interface) and back end (database) of the web tool were deployed on our university’s cloud infrastructure. Ethical approval for the study was obtained from the university's ethics committee prior to commencing the annotation process with the target group participants. All target group participants were compensated at a uniform rate of 10 euros per working hour.

The target group annotation sessions were facilitated by our Austria-based project partner. Prior to the start of each session, an educational coordinator (i.e., a proctor) supporting the target group participants read aloud the web tool instructions and consent form, both of which were written in simplified language. The coordinator also demonstrated the annotation process through example tasks to familiarize the target group participants with the procedure. Each participant was provided with a tablet and unique log-in ID. Once they accessed the web tool, they annotated independently under the coordinator's supervision. Overall, we organized 15 annotation sessions, each attended by 1-10 of our 15 target group participants. The participants had an average age of 27.4, and each was previously assessed to have a mild to moderate intellectual disability. 

\begin{figure}[htb]
    \centering
    \includegraphics[width=\linewidth]{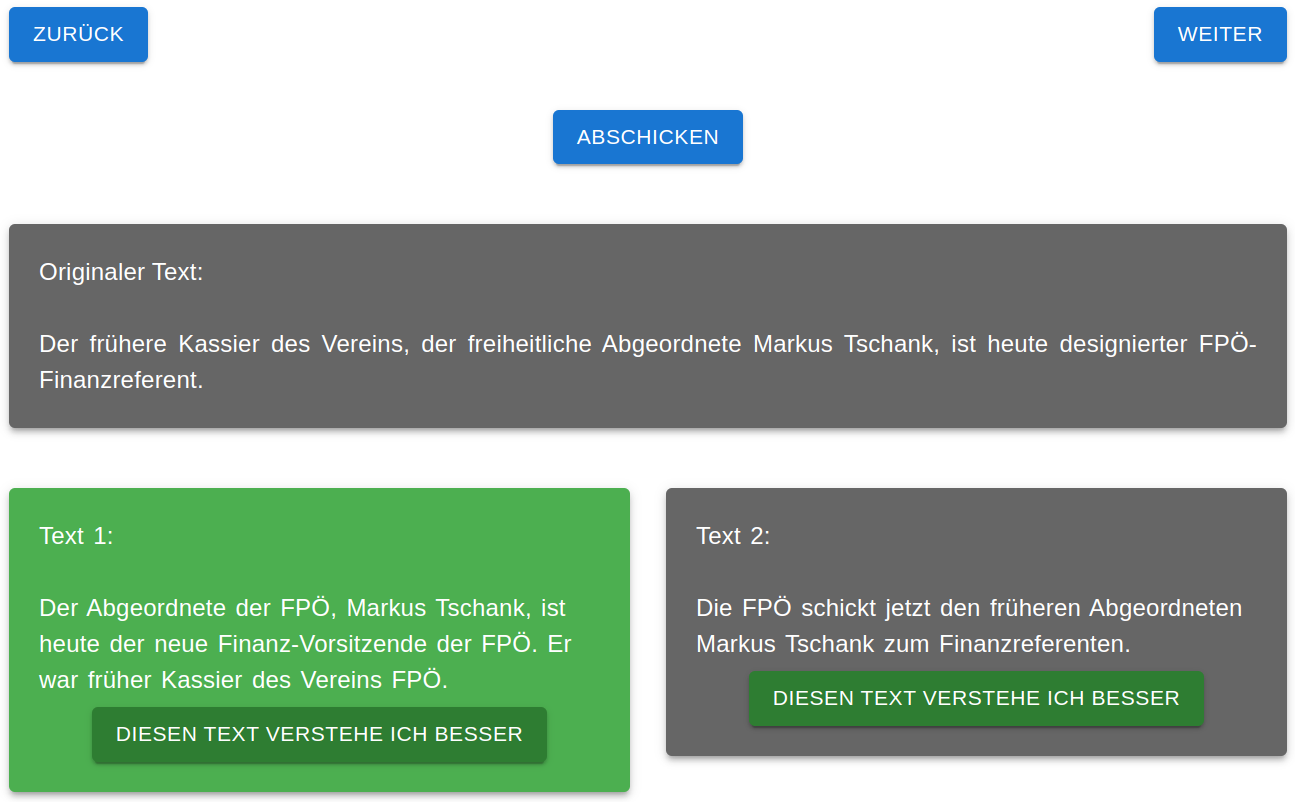}
    \caption{\textbf{A web application} was developed to collect preference annotations over paired text simplifications from both target group and expert group participants. During the annotation sessions, the expert group annotators with userIDs ea02 or ea04 were shown each pair's corresponding complex text, while those expert group annotators with user IDs ea01 or ea03 as well as all target group annotators were not shown any complex texts. As an indicator, the preferred text simplification is highlighted with a green background upon annotation. All data was hosted on our university cloud to adhere to data privacy regulations.}
    \Description{The figure shows a screenshot of a web-based annotation interface designed for pairwise preference selection between two simplified texts. At the top of the screen are three rectangular blue buttons labeled "ZURÜCK" (Back), "ABSCHICKEN" (Submit), and "WEITER" (Next), positioned on the left, center, and right respectively. Below these, a dark gray box labeled "Originaler Text" contains the original complex German sentence: “Der frühere Kassier des Vereins, der freiheitliche Abgeordnete Markus Tschank, ist heute designierter FPÖ-Finanzreferent.” Underneath, two side-by-side boxes present simplified versions of the original text, labeled “Text 1” on the left and “Text 2” on the right. Each simplified text is placed in a rounded rectangular container. Text 1's rectangular container is displayed with a green background, indicating it was selected as the preferred option, and reads: “Der Abgeordnete der FPÖ, Markus Tschank, ist heute der neue Finanz-Vorsitzende der FPÖ. Er war früher Kassier des Vereins FPÖ.” Text 2 reads: “Die FPÖ schickt jetzt den früheren Abgeordneten Markus Tschank zum Finanzreferenten,” and is shown on a dark gray background, indicating it was not selected. Both texts are accompanied by a green button below labeled “DIESEN TEXT VERSTEHE ICH BESSER” (“I understand this text better”), allowing users to register their preference.}
    \label{fig:web-tool-expert}
\end{figure}

In parallel, we recruited four native German-speaking human annotators with expertise in text simplification (the expert group) to perform the same preference annotation task on the HF4ATS-DPO dataset. To reduce inter-group bias, two of the expert group participants (denoted in our results as ea01 and ea03) were unable to see each pair's corresponding complex text, matching the target group's annotation conditions. To balance this inter-group bias reduction with a desire to leverage the expert group's professional training, we displayed corresponding complex texts to the two remaining expert group participants (ea02, ea04). This latter design is illustrated in Figure~\ref{fig:web-tool-expert}; the view for all other participants had the box containing ``Originaler Text:'' (English: ``Original Text:'') removed. No participant was made aware of the difference in annotation conditions.

Expert group participants were compensated for their annotations at an hourly rate. Ethical approval for the expert group annotations was not required. 
A detailed task instruction sheet and a tutorial video were provided to all expert group annotators prior to the kick-off of the annotation task. Overall, the expert group participants completed the annotation tasks significantly faster than most target group participants; during the final evaluation annotations, expert annotators averaged 180 pairs per hour against the target group's 60 pairs per hour.

To measure preference consistency within each annotation group and each annotator, we introduced repeated (within-annotator consistency) and shared (within-group consistency) pairs into the annotation process. We targeted a per-participant incidence of 40-45 repeated pairs and 40-45 shared pairs, in total amounting to 10\% of each expert group participant's annotations and at least 10\% of each target group participant's annotations. Sanity check pairs were randomly sorted into each participant's set and pairs' constituent simplifications were presented in random order upon each appearance. These sanity check pairs were then used to calculate intra- and inter-annotator agreement (Intra-AA and Inter-AA) to measure individual preference consistency and group-level preference consistency, respectively. Inter-AA was computed separately for the target group and the expert group. We report Intra- and Inter-AA results in Table~\ref{tab:IAA}.

\setlength{\aboverulesep}{0pt}
\setlength{\belowrulesep}{0pt}

\begin{table*}[t]
    \centering
    \begin{subtable}[t]{0.53\textwidth}
        \centering
        \caption{\textbf{Intra-annotator agreement (Intra-AA)} of target and expert group participants. We calculated Cohen's Kappa \citep{cohen1960coefficient} as the agreement score. NA marks those annotators for which Intra-AA data is not available; ta06, ta08, and ta09 data was not included in DPO post-training while ta13 and ta15 could not attend sessions with Intra-AA pairs.}
        \resizebox{0.9\textwidth}{!}{
        \begin{tabular}{ll|ll|ll|ll}
            \hline
            \multicolumn{6}{c|}{\textbf{Target}} & \multicolumn{2}{c}{\textbf{Expert}}\\
            \hline
            
            \textbf{id} & $\kappa$ & \textbf{id} & $\kappa$ & \textbf{id} & $\kappa$ & \textbf{id} & $\kappa$ \\
            \hline
            ta01 & -0.037 & ta06 & NA    & ta11 & 0.063 & ea01 & 0.420 \\
           
            ta02 & 0.040  & ta07 & -0.045 & ta12 & 0.155 & ea02 & 0.755 \\
            
            ta03 & -0.026 & ta08 & NA    & ta13 & NA & ea03 & 0.745 \\
           
            ta04 & 0.168  & ta09 & NA    & ta14 & 0.008 & ea04 & 0.376 \\
            
            ta05 & 0.300  & ta10 & 0.065 & ta15 & NA & & \\
            \hline
        \end{tabular}
        }
        \label{tab:Intra-AA}
    \end{subtable}
    \hfill
    \begin{subtable}[t]{0.43\textwidth}
        \centering
        \caption{\textbf{Inter-annotator agreement (Inter-AA)} of target and expert group participants. We calculated Krippendorff's Alpha \citep{krippendorff2004reliability} as the agreement score. We report Inter-AA scores for pairs annotated by at least four annotators, stratified by the generating SFT checkpoint.}
        \resizebox{0.9\textwidth}{!}{
        \begin{tabular}{l|c|c}
            \hline
            \multirow{2}{*}{\textbf{SFT Checkpoint}} & \multicolumn{2}{c}{$\alpha$} \\
            \cline{2-3}
            & \textbf{Target} & \textbf{Expert} \\
            \hline
            DiscoLeo-Llama-SFT-2800 & 0.019 & 0.324 \\
            Llama-SFT-2400           & 0.003 & 0.248 \\
            LeoLM-Mistral-SFT-1600   & -0.016 & 0.536 \\
            \hline
        \end{tabular}
        }
        \label{tab:Inter-AA}
    \end{subtable}
    
    \Description{Subtable (a) reports Cohen’s Kappa scores for intra-annotator agreement per participant, with NA marking missing data. Subtable (b) shows Krippendorff’s Alpha inter-annotator agreement scores for three SFT checkpoints across both target and expert groups.}

    \caption{\textbf{Annotator agreement scores} measured for target and expert group participants.}
    \label{tab:IAA}
\end{table*}

% In addition to human annotations, we incorporated a third source of preferences by leveraging another mainstream LLM—Qwen \citep{bai2023qwen, yang2024qwen2}—as a synthetic preference annotator. Specifically, we employed the pre-trained Qwen-2.5-7B-Instruct model to pair text simplifications generated by other SFT checkpoints for DPO training. The prompt used for this purpose is provided in Appendix B. Using this LLM-as-judge approach, we generated an additional 1,000 DPO preference pairs, further expanding the coverage and diversity of our dataset.

In Table~\ref{tab:dataset-stat}, we showcase the overall statistics for our HF4ATS dataset. As described above, 70\% of the 5,200 HF4ATS-SFT ($\mathcal{D}_{\text{SFT}}$) complex-simple text pairs sampled from \textsc{DEplain}-APA were used to conduct SFT. $\mathcal{D}_{\text{SFT}}$ development data was used for hyperparameter tuning and select winning SFT checkpoints for DPO post-training, while the $\mathcal{D}_{\text{SFT}}$ test data was reserved to evaluate winning DPO checkpoints against pre-DPO winning SFT checkpoints in Section~\ref{sec:results}. 

HF4ATS-DPO ($\mathcal{D}_{\text{DPO}}$), meanwhile, represents the 6,018 ATS preference pairs---3,009 unique pairs each annotated at least once by a target group participant and at least once by an expert group participant---used to train and evaluate our DPO implementation. 80\% of $\mathcal{D}_{\text{DPO}}$ pairs were used to post-train our DPO policy models, 15\% were used to select winning DPO checkpoints based on win rates, and a final 15\% were used to calculate these winning DPO checkpoints' win rates on withheld data. We also report the rate at which annotators preferred the first ATS shown (i.e., the left-hand side ATS) on the annotation tool. Due to randomization in the order of presentation, we would expect this number to be close to 50\% provided annotators adhered to task instructions.

\begin{table*}[htb]
    \centering
    \begin{tabular}{l|c|c|c|c|c|c|c|c}
     \hline
     \multirow{2}{*}{\textbf{Dataset}} & \multicolumn{3}{c|}{\textbf{\# Instances}} & \multicolumn{3}{c|}{\textbf{\# words}} & \multicolumn{2}{c}{\textbf{Pref. \% of 1st. ATS}}  \\
     \cmidrule{2-9}
     % & \textbf{Train} & \textbf{Dev} & \textbf{Test} & & \textbf{Train} & \textbf{Dev} & \textbf{Test} & & \cellcolor{gray!25} \textbf{Target} & \cellcolor{gray!25} \textbf{Expert}\\
     & \textbf{Train} & \textbf{Dev} & \textbf{Test} & \textbf{Train} & \textbf{Dev} & \textbf{Test} & \textbf{Target} & \textbf{Expert}\\
     \hline
     HF4ATS-SFT ($\mathcal{D}_{\text{SFT}}$)  &  3,600 & 800 & 800 & 252,285 & 55,208 & 55,852 & -  & - \\
     HF4ATS-DPO ($\mathcal{D}_{\text{DPO}}$) & 4,814 & 602 & 602 & 372,687 & 45,857 & 45,992 & 36.65 & 47.44 \\
     \hline
    \end{tabular}
    \vspace{1mm}
    \Description{The figure shows Table 2, titled “Overall statistics of the HF4ATS dataset.” It compares two datasets—HF4ATS-SFT and HF4ATS-DPO—used respectively for supervised fine-tuning and direct preference optimization. The table includes columns for the number of data instances and total word counts in the training, development, and test data subsets, as well as average target or expert group preference for the automatic text simplification displayed on the left-hand side of our web annotation tool (Pref. \% of 1st. ATS). In the HF4ATS-SFT dataset, there are 3,600 training instances, 800 development instances, and 800 test instances, with corresponding word counts of 252,285, 55,208, and 55,852. Preference percentages are not reported for this dataset. The HF4ATS-DPO dataset contains 4,814 training instances, 602 development instances, and 602 test instances, with 372,687, 45,857, and 45,992 words, respectively. For HF4ATS-DPO, 36.65\% of target group annotations and 47.44\% of expert group annotations selected the left-hand simplification displayed. A caption beneath the table explains that these datasets were curated for training models using either supervised fine-tuning or DPO, and that the preference percentage indicates how often the first simplification was favored, referencing Figure 3 for context.}
    \caption{\textbf{Overall statistics of the HF4ATS dataset}. We curated separate datasets for training SFT models from pre-trained LLMs and for training DPO models using preference annotations collected from either target group participants or expert group participants. Pref. \% 1st. indicates the percentage of cases in which the automatic text simplification displayed on the left of our web application (see Figure~\ref{fig:web-tool-expert}) was preferred by the respective annotator group.}
    \label{tab:dataset-stat}
\end{table*}

Note that the number of annotated preference pairs available in HF4ATS is larger than reported for $\mathcal{D}_{\text{DPO}}$ in Table~\ref{tab:dataset-stat}. This is because preference pairs annotated more than once were de-duplicated, new annotation data continued to be collected after DPO post-training was implemented, and some preference pairs were excluded from training due to grammar or content errors. The public data release contains a field indicating whether a pair annotation was included in $\mathcal{D}_{\text{DPO}}$.

\subsubsection{DPO Post-training} 
\label{sec:dpo-post-training}

\setlength{\aboverulesep}{0pt}
\setlength{\belowrulesep}{0pt}

\begin{table*}[!htb]
    \centering
    \resizebox{0.9\textwidth}{!}{
    \begin{tabular}{l|l|l|l}
     \hline
     \multirow{2.5}{*}{\textbf{Pre-trained LLMs}} & \multirow{2.5}{*}{\textbf{SFT Checkpoint}} & \multicolumn{2}{c}{\textbf{DPO Checkpoint}} \\    
     \cmidrule[0.5pt]{3-4}
    %  & &  
    % \multicolumn{1}{c}{\cellcolor{gray!20} \textbf{\rule{0pt}{2ex} Target}}  
    % & \multicolumn{1}{c|}{\cellcolor{gray!20} \textbf{\rule{0pt}{2ex} Expert}} \\
    & &  
    \multicolumn{1}{c|}{\textbf{\rule{0pt}{2ex} Target}}  
    & \multicolumn{1}{c}{\textbf{\rule{0pt}{2ex} Expert}} \\
     \hline 
     DiscoLeo-Llama-3-8B-Instruct & DiscoLeo-Llama-SFT-2800 & DiscoLeo-Llama-DPO-2160 & DiscoLeo-Llama-DPO-1080 \\
     Llama-3.1-8B-Instruct & Llama-SFT-2400 & Llama-DPO-1440 & Llama-DPO-1320 \\
     LeoLM-Mistral-7B-Chat & LeoLM-Mistral-SFT-1600 & LeoLM-Mistral-DPO-1560 & LeoLM-Mistral-DPO-2280 \\
     \hline
    \end{tabular}
    }
    \vspace{1mm}
    \Description{The figure presents Table 3, which outlines the mapping from pre-trained large language models (LLMs) to their corresponding supervised fine-tuning (SFT) and Direct Preference Optimization (DPO) checkpoints. The table consists of three main columns: Pre-trained LLMs, SFT Checkpoints, and DPO Checkpoints, with the latter divided into two subcolumns for the Target and Expert groups. The first row corresponds to DiscoLeo-Llama-3-8B-Instruct, which maps to the SFT checkpoint DiscoLeo-Llama-SFT-2800, and two separate DPO checkpoints: DiscoLeo-Llama-DPO-2160 for the target group and DiscoLeo-Llama-DPO-1080 for the expert group. The second row corresponds to Llama-3.1-8B-Instruct, with SFT checkpoint Llama-SFT-2400 and DPO checkpoints Llama-DPO-1440 (target) and Llama-DPO-1320 (expert). The third row refers to LeoLM-Mistral-7B-Chat, mapped to the SFT checkpoint LeoLM-Mistral-SFT-1600 and the DPO checkpoints LeoLM-Mistral-DPO-1560 for target and LeoLM-Mistral-DPO-2280 for expert. The caption emphasizes that DPO checkpoints were trained separately based on annotations from target and expert participants, whereas the SFT checkpoints were not group-specific.}
    \caption{\textbf{Model sequences from our pre-train $\rightarrow$ SFT $\rightarrow$ DPO pipeline} used to personalize LLM-based ATS. We reiterate that DPO checkpoints were trained separately using target and expert group annotations from HF4ATS-DPO, while SFT checkpoints were not group-specific.}
    % For the DPO stage, we trained separate DPO checkpoints using the HF4ATS-DPO-target and HF4ATS-DPO-expert datasets
    \label{tab:sft-dpo-model}
\end{table*}

Table~\ref{tab:sft-dpo-model} lists all winning model checkpoints involved in our overall training pipeline. The numbers following the SFT and DPO checkpoints indicate the number of training instances associated with each checkpoint. The DPO checkpoints refer specifically to our training variant that included all group-appropriate preference pairs in $\mathcal{D}_{\text{DPO}}$.

Starting from pre-trained LLMs, we first conducted SFT on $\mathcal{D}_{\text{SFT}}^{\text{train}}$ and selected the best-performing SFT checkpoints based on offline evaluation of $\mathcal{D}_{\text{SFT}}^{\text{dev}}$. These SFT checkpoints served as the initialization for DPO policy models and were used as frozen reference models during the DPO phase. The DPO policy models were trained on $\mathcal{D}_{\text{DPO}}^\text{train}$ using either target or expert annotations, with the best DPO checkpoints selected by offline evaluation of $\mathcal{D}_{\text{DPO}}^{\text{dev}}$ (with win rates, described in Section \ref{sec:eval}, being the most important metric). Apart from reducing the training batch size to 8 from 16, we made no changes to the SFT phase training parameters described in Appendix~\ref{sec:training-details}. Our $\beta$ parameter was to 0.1.

To study the impact of various factors on DPO post-training, we collected different subsets of preference pairs from $\mathcal{D}_{\text{DPO}}^\text{train}$ and trained DPO policy models on each. These subsets are denoted as follows:
\begin{itemize}[left=0pt]
    \item ${\text{all}}$: all preference pairs;
    \item ${\text{all}_{=}}$: all preference pairs with equal information, as indicated by pair creators;
    \item ${\text{LLM}_=}$: all preference pairs generated by the SFT checkpoint being post-trained with DPO;
    \item ${\text{max. Intra-AA}}$: all preference pairs annotated by the four target group participants or two expert group participants exhibiting the highest Intra-AA scores;
    \item ${\text{max. Inter-AA}}$: all preference pairs annotated by the four target group participants or two expert group participants exhibiting the highest Inter-AA scores.
\end{itemize}

Our final evaluation is mostly concentrated on the ``all'' subset (which is essentially responsible for the winning DPO checkpoints referenced in Table \ref{tab:sft-dpo-model}).

Given that these subsets differ in the number of training instances, one byproduct of their implementation is insight into the amount and type of preference data required to effectively personalize LLMs with 7-8 billion parameters. Such insights could inform more efficient planning of data annotation efforts involving target group participants.

\section{Evaluation}
\label{sec:eval}

\subsection{Automatic Evaluation}

Each subset of preferences listed above resulted in six DPO post-trainings, one for each combination of SFT checkpoint and annotator group. To evaluate the six winning DPO checkpoints trained on all group-appropriate preference pairs, we used greedy decoding to generate inferences for all 800 complex sentences in $\mathcal{D}_{\text{SFT}}^{\text{test}}$, calculated the reference-based metrics SARI \citep{xu2016optimizing} and BERTScore \citep{zhang2019bertscore} as well as the reference-free metric $\text{WSTF}_4$ \citep{flesch1948new}, and compared these metrics to the same metrics calculated with our winning SFT checkpoints. 

We then utilized $\mathcal{D}_{\text{DPO}}^{\text{test}}$ to calculate win rates \citep{rafailov2023direct} for all 30 winning DPO checkpoints from our trainings. Specifically, given $\mathcal{D}_{\text{DPO}}^{\text{test}} = \{ (x_i, y_i^w, y_i^l)\}_{i=1}^N$, where $x$, $y_w$, and $y_l$ denote the complex sentence, the preferred text simplification, and the dispreferred text simplification respectively, the \textbf{win rate} $W_{y_w \succ y_l}$ is defined as the proportion of preference pairs for which the DPO checkpoint assigns a higher implicit reward to the preferred text simplification than the dispreferred simplification. That is, 
\begin{align*}
    W_{y_w \succ y_l} = \frac{1}{N} \sum_{i=1}^N \mathbf{1} [ \hat{r}(x_i,y_i^w, y_i^l) > 0 ],
\end{align*}
where $\hat{r}(x,y_w, y_l)$ denotes the log-odds ratio computed with the policy model and reference model and $\mathbf{1}[\cdot]$ is the indicator function, which equals 1 if the condition holds and 0 otherwise. 
A win rate above 0.50 indicates that the DPO policy model more often assigns higher implicit rewards to human-preferred simplifications than dispreferred simplifications, thereby achieving closer alignment with human judgments of ATS quality. 

We also report the progression of average reward margins during our DPO trainings. This corresponds to the average of the implicit reward difference seen inside the indicator function.

\subsection{Human Evaluation}

We also aimed to assess whether the target group and expert group participants indeed favored text simplifications produced by models that were personalized using their own preferences as supervision signals. 

For this we calculated the DPO \textbf{supremacy score} $S_{\text{DPO} \succ \text{SFT}}$ for the six winning DPO checkpoints trained on all group-appropriate preference pairs. This score is defined as the proportion of text simplifications generated by the DPO checkpoint that human evaluators prefer over simplifications produced by the corresponding SFT checkpoint (i.e., the reference model). That is, if $(y_{\text{DPO}}^{x_i},y_{\text{SFT}}^{x_i})$ is a pairing of one DPO-checkpoint inference and one SFT-checkpoint inference, where the SFT checkpoint is the precursor of the DPO checkpoint, and $x_i \in \mathcal{D}_{\text{DPO}}$ is the corresponding complex text used to generate the two inferences, then we have
\begin{align*} 
    S_{\text{DPO} \succ \text{SFT}} = \frac{1}{N} \sum_{i=1}^N h(x_i),
\end{align*}
where
\begin{align*}
    \forall x_i \in \mathcal{D}_{\text{DPO}}^{\text{test}}, \quad h(x_i) = \begin{cases} 1, & y_{\text{DPO}}^{x_i} \succ y_{\text{SFT}}^{x_i} \\ 0,  & \text{otherwise}. \end{cases} 
\end{align*}
In this context, the successful personalization of a DPO model would thus be indicated by a supremacy score greater than 50\%.

To compute the DPO supremacy score, for every complex sentence $x$ in $\mathcal{D}_{\text{DPO}}^{\text{test}}$, we generated five text simplifications with the DPO checkpoint and five text simplifications with the corresponding SFT checkpoint using top-p sampling ($p=0.9$). We then engaged one pair creator who had previously created pairs for HF4ATS-DPO to assemble from these inferences a final set of 300 ATS pairs, 50 for each of our six winning DPO checkpoints. 

The pair creator was shown a complex sentence and a procession of possible ATS pairs in randomized order without being informed which checkpoints were responsible for each inference. The creator approved or rejected pairs based on the same criteria used during initial ATS pair creation. Only those complex sentences for which the pair creator could approve one pair for all six DPO checkpoints were included in the final evaluation round with human participants. 

We invited the four target group participants with the highest Intra-AA scores (i.e., ta04, ta05, ta10, and ta12) and all four expert annotators to take part in the final human evaluation sessions. Apart from the fact that all pairs were shared within the annotation groups (albeit displayed in randomized order), annotation conditions were the same as before. Importantly, only the 150 pairs associated with the three target-group DPO checkpoints were shown to target group annotators, and only the 150 pairs associated with the three expert-group DPO checkpoints were shown to expert group annotators. Based on pairwise choices between SFT- and DPO-checkpoint-generated text simplifications, we computed DPO supremacy scores separately for each evaluator.

\section{Results and Discussion}
\label{sec:results}

With the experiment results, we now summarize our main findings to the RQs we proposed in Section~\ref{sec:LLM-personalization} to be investigated in this study.

\subsection{Quality Assessment of Generated ATS}

\paragraph{\textbf{RQ1}.} Can DPO post-training with pairwise human preferences further improve the quality of ATS, as measured by automatic evaluation metrics?

In Table~\ref{tab:auto-eval}, we present an automatic quality assessment of ATS outputs generated by our SFT and DPO checkpoints, evaluated using both reference-based (SARI, BERTScore) and reference-free ($\text{WSTF}_4$, Win Rate) metrics.
The best performance on SARI and $\text{WSTF}_4$ was achieved with the DPO checkpoints, whereas the highest BERTScore was obtained using the SFT checkpoints.

The increased performance of DPO checkpoints measured by $\text{WSTF}_4$ show that DPO post-training benefits the ATS readability in general, regardless of the backbone LLM or preference source. Meanwhile, DPO checkpoints have consistently lower BERTScores than the corresponding SFT checkpoints across all LLMs and both preference sources, suggesting that DPO post-training has caused a certain level of semantic drift. While the effect on simplification faithfulness (as measured by SARI) is not uniform, DPO checkpoints trained with expert group preferences—such as Llama-DPO-1320 or LeoLM-Mistral-DPO-2280—tend to maintain or recover baseline SARI scores. This contrasts with DPO training on target group preferences, where decreases in simplification faithfulness were observed in two target DPO checkpoints. Importantly, expert-supervised models uniformly achieve higher win rates than their corresponding target-supervised models, emphasizing greater preference consistency among expert group annotations. This may explain the two supervision sources' differing impacts on readability and faithfulness. The findings above point to a trade-off between simplification strength and meaning preservation, with the consistency of supervision playing a key role in how well DPO models balance the two.

\begin{table*}[htb]
    \centering
    % \resizebox{\textwidth}{!}{
    \begin{tabular}{l|c|c|c|c}
    \hline
    \multirow{1}{*}{\textbf{Checkpoint}} & \multicolumn{2}{c|}{\textbf{Reference-based Metrics}} & \multicolumn{2}{c}{\textbf{Reference-free Metrics}} \\
    \hline
    %\rowcolor{gray!20} \multicolumn{6}{l}{\textbf{SFT Baselines}} \\
    \multicolumn{1}{l|}{\textbf{SFT Baselines}} & \multicolumn{1}{c|}{\textbf{SARI}} & \multicolumn{1}{c|}{\textbf{BERTScore}} & \multicolumn{1}{c|}{\textbf{$\text{WSTF}_4$}} & \multicolumn{1}{c}{\textbf{Win Rate}} \\
    \hline
    DiscoLeo-Llama-SFT-2800 & \textbf{46.22} {\small $\pm$ 13.47} & \textbf{0.9049} {\small $\pm$ 0.054} & 6.515 {\small $\pm$ 3.24} & - \\
    Llama-SFT-2400 & 45.94 {\small $\pm$ 13.52} & \textbf{0.8865} {\small $\pm$ 0.054} & 5.852 {\small $\pm$ 2.90} & - \\
    LeoLM-Mistral-SFT-1600 & 44.55 {\small $\pm$ 13.95} & \textbf{0.9054} {\small $\pm$ 0.056} & 6.207 {\small $\pm$ 3.55} & - \\
    \hline
    % \rowcolor{gray!20} 
    % \multicolumn{6}{l}{\textbf{DPO Target}} \\
    \multicolumn{1}{l|}{\textbf{DPO Target}} & \multicolumn{1}{c|}{\textbf{SARI}} & \multicolumn{1}{c|}{\textbf{BERTScore}} & \multicolumn{1}{c|}{\textbf{$\text{WSTF}_4$}} & \multicolumn{1}{c}{\textbf{Win Rate}} \\
    \hline
    DiscoLeo-Llama-DPO-2160 & \textcolor{BrickRed}{44.41} {\small $\pm$ 11.60} & \multicolumn{1}{l|}{\textcolor{BrickRed}{0.7854} {\small $\pm$ 0.081}} & \textcolor{RoyalBlue}{6.194} {\small $\pm$ 2.17} & \textcolor{RoyalBlue}{0.5211} \\
    Llama-DPO-1440 & \textcolor{RoyalBlue}{46.11} {\small $\pm$ 11.60} & \multicolumn{1}{l|}{\textcolor{BrickRed}{0.8756} {\small $\pm$ 0.055}} & \textcolor{RoyalBlue}{5.796} {\small $\pm$ 2.56} & \textcolor{RoyalBlue}{0.5145} \\
    LeoLM-Mistral-DPO-1560 & \textcolor{BrickRed}{43.73} {\small $\pm$ 13.36} & \multicolumn{1}{l|}{\textcolor{BrickRed}{0.7781} {\small $\pm$ 0.113}} & \textcolor{RoyalBlue}{5.683} {\small $\pm$ 2.80} & \textcolor{BrickRed}{0.4382} \\
    \hline
    %\rowcolor{gray!20} \multicolumn{6}{l}{\textbf{DPO Expert}} \\
    \multicolumn{1}{l|}{\textbf{DPO Expert}} & \multicolumn{1}{c|}{\textbf{SARI}} & \multicolumn{1}{c|}{\textbf{BERTScore}} & \multicolumn{1}{c|}{\textbf{$\text{WSTF}_4$}} & \multicolumn{1}{c}{\textbf{Win Rate}} \\
    \hline
    DiscoLeo-Llama-DPO-1080 & \textcolor{BrickRed}{42.50} {\small $\pm$ 13.28} & \multicolumn{1}{l|}{\textcolor{BrickRed}{0.7814} {\small $\pm$ 0.059}} & \textbf{\textcolor{RoyalBlue}{4.031}} {\small $\pm$ 2.68} & \textbf{\textcolor{RoyalBlue}{0.6118}} \\
    Llama-DPO-1320 & \textbf{\textcolor{RoyalBlue}{46.45}} {\small $\pm$ 12.25} & \multicolumn{1}{l|}{\textcolor{BrickRed}{0.8441} {\small $\pm$ 0.052}} & \textbf{\textcolor{RoyalBlue}{4.676}} {\small $\pm$ 2.59} & \textbf{\textcolor{RoyalBlue}{0.6099}} \\
    LeoLM-Mistral-DPO-2280 & \textbf{\textcolor{RoyalBlue}{44.92}} {\small $\pm$ 14.03} & \multicolumn{1}{l|}{\textcolor{BrickRed}{0.8340} {\small $\pm$ 0.082}} & \textbf{\textcolor{RoyalBlue}{4.802}} {\small $\pm$ 2.94} & \textbf{\textcolor{RoyalBlue}{0.6118}} \\
    \hline
    \end{tabular}
    %}
    \vspace{1mm}
    \caption{\textbf{Automatic assessment of ATS quality} for inferences from both SFT and DPO checkpoints on the SFT test set $\mathcal{D}_{\text{SFT}}^{\text{test}}$ (columns 1-3) and DPO test set $\mathcal{D}_{\text{DPO}}^{\text{test}}$ (column 4). We report the mean scores and standard deviations (in brackets) for each metric.
    %Performance improvements of DPO checkpoints over their SFT counterparts are highlighted in \textcolor{RoyalBlue}{blue}, and performance declines in \textcolor{BrickRed}{red}. 
    The same color scheme is used for win rates (values above vs. below 0.50). DPO checkpoints in the table were trained with all annotated data for the given supervision source (expert or target). The relatively high standard deviations for SARI reflect the fact that only one reference per test instance is available. Win rates are computed based on annotator preferences within the respective supervisory groups (target or expert). %For the first three metrics, we mark the best performance across models in \textbf{bold font}.
    }
    \Description{The figure presents Table 4, titled “Automated, quantitative assessment of ATS quality,” which reports evaluation scores for different model checkpoints across both reference-based and reference-free metrics. The table is structured into three main model categories: “SFT Baselines,” “DPO Target,” and “DPO Expert.” Each row lists a specific checkpoint and its performance across four metrics: SARI, BERTScore, WSTF4, and Win Rate. The SFT Baselines category includes DiscoLeo-Llama-SFT-2800, Llama-SFT-2400, and LeoLM-Mistral-SFT-1600, all evaluated on the supervised fine-tuning test set. These SFT model rows report SARI scores between 44.55 and 46.22, BERTScore values near 0.90, and WSTF4 scores ranging from 5.85 to 6.52. Win rates are not applicable for these baselines and are left blank. The DPO Target section evaluates DiscoLeo-Llama-DPO-2160, Llama-DPO-1440, and LeoLM-Mistral-DPO-1560. Among these, Llama-DPO-1440 shows strong performance, with a SARI score of 46.11, BERTScore of 0.8756, and a win rate of 0.5145. The LeoLM checkpoint underperforms, with several metrics highlighted in red, such as a SARI of 43.73 and a win rate of 0.4382. By contrast, the DPO Expert section, which includes DiscoLeo-Llama-DPO-1080, Llama-DPO-1320, and LeoLM-Mistral-DPO-2280, shows consistently higher win rates of approximately 0.61 and improved SARI and WSTF4 scores in most cases. Metrics that show improvement over the SFT baselines are marked in green, while deterioration is marked in red. The table caption explains that the scores represent means and standard deviations, with SARI variability attributed to single-reference evaluation. Win rates are computed based on preferences within target or expert groups. The color coding visually distinguishes gains (green) and drops (red) in performance, particularly in the DPO checkpoints compared to their SFT counterparts. Win rates are computed based on annotator preferences within the respective supervisory groups (target or expert).}
    \label{tab:auto-eval}
\end{table*}

Recent studies have revealed several core limitations of standard DPO post-training. These include a tendency to overfit to sparse or noisy preference signals \citep{fisch2024robust}, catastrophic forgetting in continual learning settings \citep{qi2024online}, and the potential to undermine generalization and robustness in LLMs \citep{hu2024new}. In our study, these limitations may help explain the observed decline in DPO models' faithfulness with respect to the \textsc{DEplain} data in $\mathcal{D}_{\text{SFT}}^{\text{test}}$. Nonetheless, our results highlight the critical role of preference consistency in the effectiveness of DPO for personalized ATS modeling. The win rates in Table~\ref{tab:auto-eval} as well as Inter- and Intra-AA scores in Table~\ref{tab:IAA} indicate that target group preferences are more diverse or inconsistent than expert group preferences. It might be the case that offline LLM alignment methods such as DPO, which lack explicit reward modeling, are suboptimal for capturing nuanced preferences over text simplifications when trained with such data.

\paragraph{\textbf{Findings}.} Overall, DPO post-training improves ATS in terms of text readability (as evidenced by $\text{WSTF}_4$ scores) and, in some cases, N-gram preservation (as evidenced by SARI scores). However, despite evidence that expert group supervision signals can result in closer alignment of their preferences, models trained on either expert or target group supervision exhibit loss in meaning preservation (as evidenced by BERTScore). The scale of these effects depends on the backbone model—Llama-3.1-8B-Instruct, DiscoLeo-Llama-3-8B-Instruct, or LeoLM-Mistral-7B-Chat—and the source of the supervision signals, i.e., target group participants vs. expert group participants. 

\subsection{Impact of Individual Factors on DPO Post-training}

\paragraph{\textbf{RQ2}.} To what extent do factors as preference source, information equality, and generalization of LLMs influence the effectiveness of DPO post-training?

\begin{table*}[htb]
    \centering
    %\resizebox{\textwidth}{!}{
    \begin{tabular}{l|l|l|l|l|l}
    \hline
    \textbf{DPO Checkpoint}  & \textbf{all} & ${\textbf{all}_{=}}$ & ${\textbf{LLM}_=}$ & ${\textbf{max. Intra-AA}}$ & ${\textbf{max. Inter-AA}}$ 
    %& $(x,y_w,y_l)_{\text{Train$_{All_{ex}}$ $\rightarrow$ Test$_{Intra-AA_{tg}}$}}$ 
    \\
     \hline
     %\rowcolor{gray!20} \textbf{Target} & \textbf{Baseline} & &  & &  \\
     \multicolumn{1}{l|}{\textbf{Target}} & \multicolumn{1}{l|}{\textbf{Baseline}} & \multicolumn{4}{c}{\textbf{Subsets of HF4ATS-DPO training data}} \\
     \hline
     DiscoLeo-Llama-DPO & 0.5211 & 0.4708 {\small(\textcolor{BrickRed}{9.65\%} $\downarrow$)} & 0.4861 {\small(\textcolor{BrickRed}{6.72\%} $\downarrow$)} & 0.5078 {\small(\textcolor{BrickRed}{2.55\%} $\downarrow$)} & \textbf{0.5431} {\small(\textcolor{RoyalBlue}{4.22\%} $\uparrow$)} \\ % & 0.4943 \\
     Llama-DPO & 0.5145  & 0.4833 {\small(\textcolor{BrickRed}{6.06\%} $\downarrow$)} & 0.5385 {\small(\textcolor{RoyalBlue}{4.66\%} $\uparrow$)} & \textbf{0.6094} {\small(\textcolor{RoyalBlue}{18.45\%} $\uparrow$)} & 0.5153 {\small(\textcolor{RoyalBlue}{0.16\%} $\uparrow$)} \\ % & 0.5052 \\
     LeoLM-Mistral-DPO & 0.4382 & 0.4625 {\small(\textcolor{RoyalBlue}{5.55\%} $\uparrow$)} & 0.4848 {\small(\textcolor{RoyalBlue}{10.63\%} $\uparrow$)} & \textbf{0.5781} {\small(\textcolor{RoyalBlue}{31.93\%} $\uparrow$)} & 0.4917 {\small(\textcolor{RoyalBlue}{12.21\%} $\uparrow$)} \\ % & 0.5083\\
     \hline
     %\rowcolor{gray!20} \textbf{Expert} & \textbf{Baseline} & & & &  \\
     \multicolumn{1}{l|}{\textbf{Expert}} & \multicolumn{1}{l|}{\textbf{Baseline}} & \multicolumn{4}{c}{\textbf{Subsets of HF4ATS-DPO training data}} \\
     \hline
     DiscoLeo-Llama-DPO & 0.6118 & 0.6333 {\small(\textcolor{RoyalBlue}{3.51\%} $\uparrow$)} & 0.5111 {\small(\textcolor{BrickRed}{16.46\%} $\downarrow$)} & 0.6118 {\small(0.00\% $=$)} & \textbf{0.6438} {\small(\textcolor{RoyalBlue}{5.23\%} $\uparrow$)} \\ % & -\\
     Llama-DPO & 0.6099 & 0.5833 {\small(\textcolor{BrickRed}{4.36\%} $\downarrow$)} & \textbf{0.6538} {\small(\textcolor{RoyalBlue}{7.20\%} $\uparrow$)} & 0.6382 {\small(\textcolor{RoyalBlue}{4.64\%} $\uparrow$)} & 0.6313 {\small(\textcolor{RoyalBlue}{3.51\%} $\uparrow$)} \\ % & -\\
     LeoLM-Mistral-DPO & 0.6118 & 0.6125 {\small(\textcolor{RoyalBlue}{0.11\%} $\uparrow$)} & 0.5871 {\small(\textcolor{BrickRed}{4.04\%} $\downarrow$)} & \textbf{0.6776} {\small(\textcolor{RoyalBlue}{10.76\%} $\uparrow$)} & 0.5625 {\small(\textcolor{BrickRed}{8.06\%} $\downarrow$)} \\ % & -\\
     \hline
    \end{tabular}
    \vspace{1mm}
    %}
    \caption{\textbf{Win rates ($W_{y_w \succ y_l}$) on corresponding $\mathcal{D}_{\text{DPO}}^{\text{test}}$ subsets} for the winning DPO checkpoints trained on different subsets of $\mathcal{D}_{\text{DPO}}^{\text{train}}$, as described in Section~\ref{sec:dpo-post-training}. Note that the number of training instances is not uniform across the reported DPO checkpoints. To avoid potential confusion, we omit the number of training instances per subset. For each subset, we saved a DPO checkpoint every 120 training instances and retained the checkpoint with the highest win rate on the corresponding $\mathcal{D}_{\text{DPO}}^{\text{dev}}$ subset. The performance changes in brackets indicate differences relative to the results in the second column \textit{all} (i.e., the checkpoints indicated by the circles ($\circ$) in Figure~\ref{fig:win-rate-data-ablation}). %Relative to the DPO checkpoint trained on all pairs, we highlight performance increases in \textcolor{RoyalBlue}{blue} and decreases in \textcolor{BrickRed}{red}. For each model, we mark the best performance observed on the particular training subset in \textbf{bold font}.
    }
    \Description{The figure shows Table 5, titled “Win rates ($W_{y_w \succ y_l}$) on corresponding $\mathcal{D}_{\text{DPO}}^{\text{test}}$ subsets,” which compares the win rates of various DPO checkpoints trained on different subsets of the training data. The table is organized into two main sections: “Target” and “Expert,” each listing three DPO checkpoints: DiscoLeo-Llama-DPO, Llama-DPO, and LeoLM-Mistral-DPO. The columns correspond to different preference pair subsets used to fine-tune the models: “all,” “all=,” “LLM=,” “max. Intra-AA,” and “max. Inter-AA.” Each cell contains the win rate on the test set corresponding to the given preference pair subset, with results from the “all” column used as the baseline for comparison. The values in parentheses indicate the relative change compared to that baseline: increases are highlighted in green with upward arrows, while decreases are in red with downward arrows. For example, in the “Target” group, the LeoLM-Mistral-DPO checkpoint trained on the “LLM=” subset shows a win rate of 0.4848, which is 10.63\% higher than its performance on “all.” By contrast, the “Target” group DiscoLeo-Llama-DPO model trained on the same subset drops by 6.72\%. In the “Expert” group, LeoLM-Mistral-DPO trained on “max. Intra-AA” achieves the win rate at 0.6776, which is a 10.76\% improvement over its baseline and the highest win rate in the table. The caption clarifies that the win rates were computed for checkpoints trained on individual preference pair subsets as described in Section 3.2.3. Checkpoints were saved every 120 training instances per subset, and the highest-performing checkpoint on the development set during training was retained for inclusion in this table. This table allows direct comparison of how data quality and agreement measures influence test performance under DPO training.}
    \label{tab:dpo-win-rates}
\end{table*}

Table~\ref{tab:dpo-win-rates} presents the win rates of DPO checkpoints trained on different subsets of $\mathcal{D}_{\text{DPO}}^{\text{train}}$, retained because they exhibited the maximum win rate on their subset of $\mathcal{D}_{\text{DPO}}^{\text{dev}}$ during training, and evaluated on corresponding subsets of $\mathcal{D}_{\text{DPO}}^{\text{test}}$. We observe that DPO models trained on subsets reflecting higher consistency of human preferences, such as those with maximized Intra- or Inter-AA scores, leads to an almost uniform improvement in win rates across all models and both preference sources. For example, the target group LeoLM-Mistral-DPO model achieves a substantial 31.93\% boost to win rate for training on the Intra-AA subset relative to the baseline DPO training on all preference pairs. Meanwhile, subsets focused on human perception or model involvement variables, such as pairs labeled with information equity ($\text{all}_=$) or pairs with the same LLM backbone as the model being post-trained ($\text{LLM}_=$), exhibit a less consistent and frequently negative impact on performance. For instance, the win rate of the target group DiscoLeo-Llama-DPO model drops by 9.65\% under the $\text{all}_=$ subset. Every DPO model trained with expert group supervision has a higher win rate than the target-group-supervised model with the same backbone and training pair subset. Additionally, no expert-group DPO model's win rate dips below 0.50—a threshold indicating a DPO-induced drift \textit{away} from human preferences—and win rate shifts across the different training pair subsets tend to be lower for the expert-group models than the target-group models. These results suggest that preference consistency is a stronger driver of DPO effectiveness than model-specific or perception-related signals, particularly when the goal is to robustly model human preferences. We emphasize that the benefits of preference consistency are not limited to expert participants; consider that the aforementioned win rate boost for LeoLM-Mistral-DPO under maximum Intra-AA target group supervision represents the largest boost across all of our experiments.

\begin{figure*}[!htb]
    \centering
    \includegraphics[width=\textwidth]{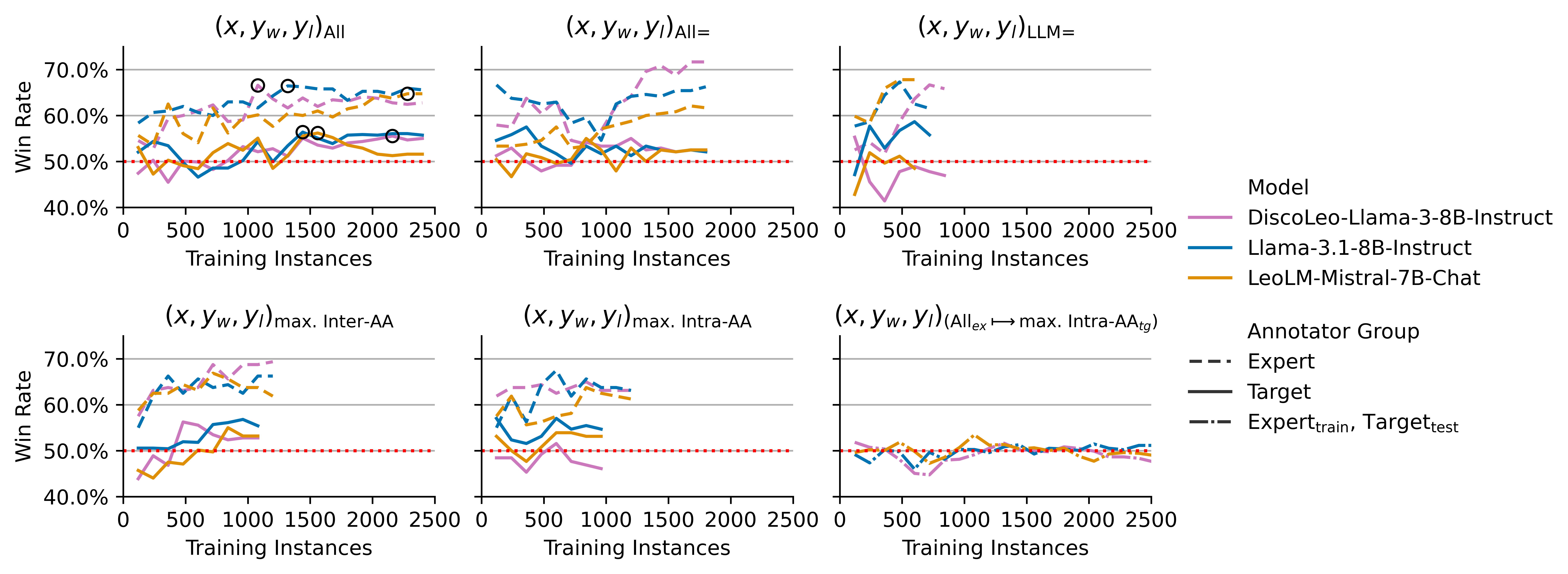}
    \caption{Win rates on development sets during training across different subsets of the HF4ATS-DPO training data. Development sets amount to 10\% of maximum training instances in all cases. Circles ($\circ$) in the upper left figure indicate the DPO checkpoint with the highest win rate on $\mathcal{D}_{\text{DPO}}^{\text{dev}}$. The bottom-right panel shows results from training on all expert group data and testing on target group data for the four target annotators with the highest Intra-AA scores.}
    \Description{The figure presents 6 line plots arranged in 2 rows and 3 columns, illustrating win rates on development sets during training across various subsets of the HF4ATS-DPO training data. The x-axis in all plots represents the number of training instances, ranging from 0 to 2500, while the y-axis shows the win rate, ranging from 40\% to 70\%. Each subplot corresponds to a different data selection strategy, with titles indicating the subset type. The top row contains results for the full dataset, a subset excluding preference pairs with unequal levels of information, and a subset where the preference pairs were generated only by the LLM being post-trained. The bottom row includes a subset with pairs annotated by the 2 expert annotators or 4 target annotators with the highest inter-annotator agreement, a subset with pairs annotated by the 2 expert annotators or 4 target annotators with the highest intra-annotator agreement, and a cross-group setup where models are trained on all expert data and tested on data from the 4 target group annotators with the highest intra-annotator agreement. 3 model variants are compared: DiscoLeo-Llama-3-8B-Instruct, Llama-3.1-8B-Instruct, and LeoLM-Mistral-7B-Chat. Apart from the final subplot in the second row, each model is trained and evaluated on each preference pair subset using expert annotations or target annotations. In the final plot, each model is trained with expert data and evaluated with target data. In the top-left plot, several performance peaks are marked with open circles, indicating the DPO checkpoints with the highest win rates on the corresponding development sets. A horizontal line marks the 50\% baseline across all plots, indicating the threshold for better-than-random performance. The plots reflect varying learning dynamics depending on data subset, model, and annotator group.}
    \label{fig:win-rate-data-ablation}
\end{figure*}

\begin{figure*}[htb]
    \centering
    \includegraphics[width=\textwidth]{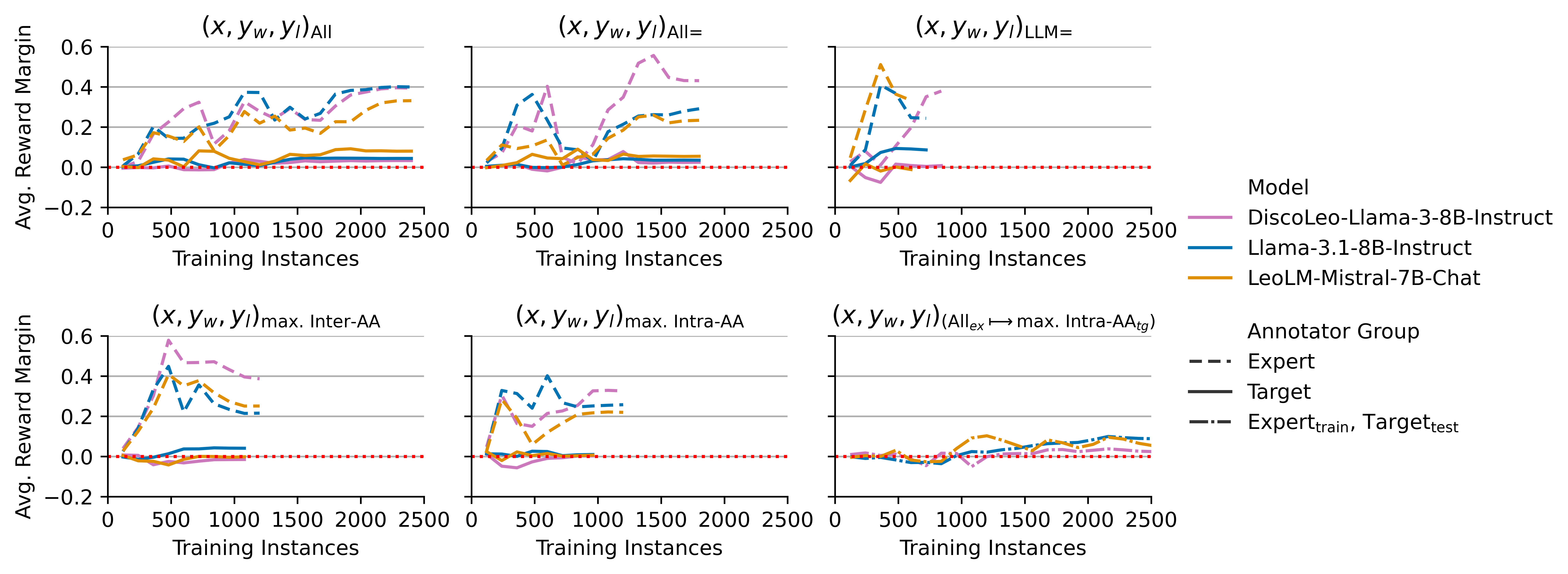}
    \caption{\textbf{Average reward margins with respect to the number of training instances} from different subsets of the HF4ATS-DPO training data, evaluated on the corresponding development sets.}
    \Description{The figure displays six line plots arranged in two rows and three columns, showing average reward margins on development sets during training across subsets of the HF4ATS-DPO training data. The x-axis in all plots represents the number of training instances, ranging from 0 to 2500, while the y-axis shows the average reward margin, ranging from -0.2 to 0.6. Each subplot corresponds to a different data selection strategy, with titles indicating the subset type. The top row contains results for the full dataset, a subset excluding pairs with unequal levels of information, and a subset where the pairs were generated only by the LLM being post-trained. The bottom row includes a subset with pairs annotated by the 2 expert annotators or 4 target annotators with the highest inter-annotator agreement, a subset with pairs annotated by the 2 expert annotators or 4 target annotators with the highest intra-annotator agreement, and a cross-group setup where models are trained on all expert data and tested on data from the 4 target group annotators with the highest intra-annotator agreement. Three model variants are compared: DiscoLeo-Llama-3-8B-Instruct, Llama-3.1-8B-Instruct, and LeoLM-Mistral-7B-Chat. Apart from the final subplot in the second row, each model is trained and evaluated on each preference pair subset using expert annotations or target annotations. In the final plot, each model is trained with expert data and evaluated with target data. A horizontal line is drawn at zero to indicate the baseline. The plots reveal that reward margins tend to increase in early training and plateau or vary depending on the model, subset, and annotator group. The clearest improvements are seen in the top-left and bottom-left subplots, where all models show positive margins, especially under expert annotations. In contrast, the bottom-right subplot shows nearly flat lines around the zero baseline, indicating minimal margin gain in the expert-to-target transfer scenario.}
    \label{fig:reward-margin-dev}
\end{figure*}

To examine how different LLMs behave during DPO post-training, at every 120 training instances we visualize win rates and implicit reward margins for $\mathcal{D}_{\text{DPO}}^{\text{dev}}$ pairs in Figures ~\ref{fig:win-rate-data-ablation} and ~\ref{fig:reward-margin-dev} respectively. In the lower-right panels of both figures, we also include results from training on all \textit{expert} group annotations and evaluating on \textit{target} group annotations from the maximum Intra-AA $\mathcal{D}_{\text{DPO}}^{\text{dev}}$ subset. This setup allows us to assess whether models trained with expert preferences—which are often much easier to collect in practice—can effectively cater to the preferences of target group persons. 

We observe that across most settings, win rates on expert preferences (dashed lines) exhibit more stable and consistent improvements as the number of training instances increases. 
Win rates on target preferences (solid lines), meanwhile, stabilize at lower levels and only slightly above the 50\% border line, likely reflecting the noisier or less consistent decision-making process among target group participants. The relatively small implicit reward margins for target group trainings suggest target group hesitation in distinguishing between preferred and dispreferred ATS outputs may have contributed to this inconsistency. Notably, in the lower-right panel, where models are trained on expert group preferences and evaluated on the most consistent subset of target group preferences, the win rates oscillate around the 50\% threshold, suggesting limited generalization from expert preferences to target preferences. These findings underscore the importance of both preference quality and consistency in DPO post-training outcomes and suggest that DPO may not be the most effective LLM alignment method when the goal is to personalize ATS systems for target groups persons with diverse subjective preferences.

\paragraph{\textbf{Findings}.} Factors that directly reflect the consistency of human preference signals—such as intra- or inter-annotator agreement—tend to have a more positive and reliable impact on win rates. By contrast, human-perceived information equity levels in training pairs or consistency between the model being post-trained and the pair-generating model show less prominent influence, suggesting they are less reliable factors of improved preference alignment. 

\subsection{Personalization Success Rates of Target and Expert Models}

\paragraph{\textbf{RQ3}.} Despite these challenges, can DPO post-training ultimately enable successful group-level personalization of ATS models?

\begin{table*}[htb]
    \centering
    \begin{tabular}{l|l|c|c|c|c}
    \hline
    \multirow{1}{*}{\textbf{SFT Checkpoint}} & \multirow{1}{*}{\textbf{DPO Checkpoint}} & \multicolumn{4}{c}{\textbf{DPO Supremacy Score}}  \\
    \hline
    %\rowcolor{gray!20} \textbf{Baseline} & \textbf{Target} & & & &   \\
    \textbf{Baseline} & \textbf{Target} & \multicolumn{1}{r|}{ta04} & \multicolumn{1}{r|}{ta05} & \multicolumn{1}{r|}{ta10} & \multicolumn{1}{r}{ta12}  \\
    \hline
    DiscoLeo-Llama-SFT-2800 & DiscoLeo-Llama-DPO-2160 & \textcolor{BrickRed}{0.36} & \textcolor{BrickRed}{0.40} &\textcolor{RoyalBlue}{0.56}& \textcolor{BrickRed}{0.46}   \\
    Llama-SFT-2400 & Llama-DPO-1440 & \textcolor{BrickRed}{0.40} & \textcolor{BrickRed}{0.30} & \textcolor{BrickRed}{0.38} & 0.50 \\
    LeoLM-Mistral-SFT-1600 & LeoLM-Mistral-DPO-1560 & \textcolor{BrickRed}{0.42} & \textcolor{BrickRed}{0.48} & \textcolor{RoyalBlue}{0.58}& \textcolor{BrickRed}{0.40} \\
    \hline
    \multirow{1}{*}{\textbf{SFT Checkpoint}} & \multirow{1}{*}{\textbf{DPO Checkpoint}} & \multicolumn{4}{c}{\textbf{DPO Supremacy Score}}  \\
     \hline
    %\rowcolor{gray!20} \textbf{Baseline} & \textbf{Expert} & & & & \\
    \textbf{Baseline} & \textbf{Expert} & \multicolumn{1}{r|}{ea01} & \multicolumn{1}{r|}{ea02} & \multicolumn{1}{r|}{ea03} & \multicolumn{1}{r}{ea04} \\
    \hline
    DiscoLeo-Llama-SFT-2800 & $\text{DiscoLeo-Llama-DPO-1080}^{*, **}$ & \textcolor{RoyalBlue}{0.74} & \textcolor{BrickRed}{0.46} & \textcolor{RoyalBlue}{0.68} & \textcolor{RoyalBlue}{0.72} \\
    Llama-SFT-2400 & Llama-DPO-1320 & \textcolor{RoyalBlue}{0.60} & \textcolor{BrickRed}{0.30} & \textcolor{RoyalBlue}{0.54} & \textcolor{RoyalBlue}{0.56} \\
    LeoLM-Mistral-SFT-1600 & $\text{LeoLM-Mistral-DPO-2280}^{*}$ & \textcolor{RoyalBlue}{0.68} & \textcolor{BrickRed}{0.44} & \textcolor{RoyalBlue}{0.52} & \textcolor{RoyalBlue}{0.56}\\
    \hline
    \end{tabular}
    \vspace{1mm}
    \Description{Table 6, titled “DPO supremacy scores measured with the best SFT and DPO checkpoints by LLM backbone.” The table is divided into two sections: the top section reports results for the Target group of evaluators and the bottom section for the Expert group. Each section compares three model families—DiscoLeo-Llama, Llama, and LeoLM-Mistral—across their corresponding SFT and DPO checkpoints. Each row lists a model's SFT checkpoint (under the “SFT Checkpoint” column), its corresponding DPO checkpoint (under “DPO Checkpoint”), and a set of DPO Supremacy Scores, shown across individual evaluators. For the Target group, the scores are reported for annotators ta04, ta05, ta10, and ta12; for the Expert group, the scores are reported for annotators ea01, ea02, ea03, and ea04. These scores indicate the proportion of cases where the DPO-generated simplification was preferred over the SFT version, based on human judgment. In the Expert section, group-level statistical significance is marked using asterisks. One asterisk (*) indicates significance when preferences from annotator ea02 are excluded, while two asterisks (**) indicate significance when ea02 preferences are included. For example, the DiscoLeo-Llama-DPO-1080 checkpoint shows the highest supremacy score (0.74) for ea01 and is marked with two asterisks, denoting statistically significant superiority over the SFT baseline. Similarly, LeoLM-Mistral-DPO-2280 is marked with one asterisk, indicating significance under exclusion of ea02 preferences. No checkpoint is marked as significant for the Target group of evaluators.}
    \caption{\textbf{DPO supremacy scores measured with the best SFT and DPO checkpoints by LLM backbone.} We report the average proportion of text simplifications produced by DPO checkpoints that are preferred by human evaluators over simplifications produced by corresponding SFT checkpoints. For the target evaluator group, we report human evaluation results based on the four evaluators with the highest Intra-AA scores during the preference annotation stage. 
    %We highlight DPO supremacy scores in \textcolor{RoyalBlue}{blue} if they are above 0.5 and in \textcolor{BrickRed}{red} if they are below 0.5. 
    Asterisks indicate whether a group-level binomial test on evaluation preference with majority voting indicates DPO supremacy for a given model at the 0.05 significance level. One asterisk indicates significance when ea02 preferences are excluded and two asterisks indicate significance when ea02 preferences are included.}
    \label{tab:sft_vs_dpo}
\end{table*}

Table~\ref{tab:sft_vs_dpo} presents DPO supremacy scores for our six winning DPO checkpoints at the individual annotator level. We find that three of four target annotators have a clear preference for SFT model inferences over DPO model inferences. Furthermore, there is some consistency among target evaluators: ta04 and ta05 are most forgiving to LeoLM-Mistral-DPO-1560, ta12 is most forgiving to Llama-DPO-1440, and ta10 is the only annotator to indicate preference for DPO inferences. In contrast to the target evaluators, three of four expert annotators exhibit a preference for DPO checkpoint inferences across all model backbones, and DiscoLeo-Llama-DPO-1080 performs best relative to SFT for all four annotators, typically by a sizable margin. 

To verify group-level personalization, we conducted one-sided binomial tests\footnote{We used the Python package \texttt{SciPy}, BSD-3-Clause license, available at \url{https://github.com/scipy/scipy}, to run the binomial tests.} for each model at the evaluation group level. Assuming each pair was evaluated independently, we defined the group-level preference for each test pair as the majority vote among the evaluators (tied pairs were assigned randomly). Our goal was to determine whether, across all 50 DPO supremacy test pairs, there was a statistically significant collective preference for ATS outputs generated by the DPO checkpoints. The asterisks in Table~\ref{tab:sft_vs_dpo} indicate which models had a group-level DPO supremacy greater than 0.50 with a $p$-value less than 0.05. For the expert group, we indicate results for tests both including and excluding the outlier ea02. 

Among expert evaluators, the two German-tuned backbones produced DPO checkpoints with DPO supremacy above 0.50 at the group level. No target group DPO checkpoint demonstrates DPO supremacy with statistical significance. The greater preference consistency among expert evaluators that manifested in Table \ref{tab:IAA} as well as Figures \ref{fig:win-rate-data-ablation} and \ref{fig:reward-margin-dev} evidently led to a DPO post-training that was able to anchor collective expert group preferences to a certain degree. Whether due to greater diversity in innate preferences, lack of simplification training, or barriers to annotation task adherence, collective target group preferences were not reflected by DPO post-trainings in the same way.

It should also be noted that unlike for the expert group checkpoints, the target group DPO checkpoints in Table \ref{tab:sft_vs_dpo} were trained on data which included annotations from eight additional target users with lower intra-annotation agreement than ta04, ta05, ta10, and ta12. It is therefore possible that human evaluation of a DPO checkpoint trained on a different subset of our target group preference pairs would better capture group-level preferences.

Given the impact seen in Table \ref{tab:sft_vs_dpo} of target group preference inconsistency, we now reflect on the effectiveness of DPO for ATS preference alignment for the target group persons. During the HF4ATS-DPO annotation stage, our target group annotators occasionally encountered pairs they had trouble annotating due to the similarity or shared suitability of the two ATS options. DPO, with its contrastive, distributional learning objective, is capable of exploiting supervision from these low-signal pairs provided annotations are gathered at scale. However, gathering data at scale is a high-resource endeavor that is uniquely encumbered in accessibility settings by the need to manage the cognitive load imposed on target group persons. This means the recurring request for them to read and compare competing simplifications may work against the need for numerous annotations. 

As an alternative framework, we propose exploring preference alignment methods that reduce the cognitive load on annotators, namely KTO \cite{ethayarajh2024model}. Target annotators would only need to label a standalone simplification as desirable or undesirable to implement KTO, thus removing comparison entirely. Separately, ATS preference alignment would likely benefit from KTO's ability to better incorporate loss aversion and other human cognitive biases due to its inference utility maximization objective. This suggests some dividends could be achieved by simply splitting HF4ATS-DPO preference pairs and using the restructured data to implement KTO, an approach we leave for future research. 

\paragraph{\textbf{Findings}.} Personalization under DPO can be successful at both the individual level and group level using group-wide preferences. That said, outcomes are sensitive to barriers to preference consistency.

\section{Conclusion and Future Work}
\label{sec:conclusion}

In this work, we studied the effectiveness of direct preference optimization (DPO) for personalizing LLM-based automatic text simplification (ATS) models to better reflect the preferences of individuals with intellectual disabilities. To enable this, we developed a lightweight and accessible web application for collecting pairwise human preferences from both target users and expert participants.
We introduced HF4ATS, the first and largest German-language ATS dataset combining preference annotations from both target and expert group. Using a standard pre-train $\rightarrow$ SFT $\rightarrow$ DPO pipeline, we trained and analyzed models on various subsets of this dataset, systematically investigating how preference consistency, preference source, and LLM engagement impact personalization outcomes.

Our findings highlight a fundamental challenge in aligning LLMs with human preferences: methods like DPO rely on consistency in supervision signals that is often difficult to achieve, particularly when preference data is collected from persons with intellectual disabilities. This limitation calls for future research in two interconnected directions. First, we need improved HCI approaches for eliciting preferences, including adaptive and accessible tools that support users in providing consistent feedback. Second, there is a pressing need to develop alignment objectives that are robust to uncertainty and variability in human preferences.
Key directions for investigation include: (1) identifying the most effective RLHF strategies for such personalization; (2) designing human feedback formats that more faithfully reflect human decision-making processes; and (3) determining whether sophisticated reward models are still needed to actively capture the complex utility structures underlying persons’ choices. Addressing these questions will likely require interdisciplinary insights, drawing from reinforcement learning, psycholinguistics, and cognitive science.

In future work, we will explore alternative RLHF techniques and lightweight personalization strategies that leverage small but high-quality human preference data. More broadly, we advocate for inclusive AI development that centers the voices of persons with disabilities, not merely as end-users or evaluators, but as active co-creators throughout the research and implementation process.

\section*{Acknowledgment}

This work was funded by the Swiss Innovation Agency (Innosuisse) Flagship Inclusive Information and Communication Technologies (IICT) under grant agreement PFFS-21-47. We sincerely thank all study participants, especially those from the target group. 

\bibliographystyle{ACM-Reference-Format}
%\bibliography{sample-base}
\bibliography{ref}


\newpage

\onecolumn
\appendix
\section{Prompt Templates Used for Constructing LLM Inputs}
\label{sec:prompts}

Table~\ref{tab:prompts} lists the prompt templates used in SFT and DPO, optimized by an text simplification expert who was not involved in data annotation. We randomly sampled from this prompt bank to increase the diversity of ATS generation.

\begin{table}[htb]
    \centering
    \small
    %\resizebox{\textwidth}{!}{
    \begin{tabular}{c|p{11cm}|c}
    \hline
    \textbf{No.} & \textbf{Prompt} & \textbf{Phase} \\
    \hline
     \multirow{2}{*}{1} & Schreibe den folgenden Satz in Leichter Sprache um: <\texttt{complex\_sentence}>. Bitte gib nur eine Vereinfachung an, ohne Einleitung, Alternativen oder Kommentare. & \multirow{2}{*}{SFT + DPO} \\
     \hline
     \multirow{3}{*}{2} & Vereinfache den folgenden Satz, sodass Menschen mit kognitiver Beeintr\"achtigung den vereinfachten Satz verstehen k\"onnen: <\texttt{complex\_sentence}>. Bitte gib nur die Vereinfachung an, ohne Einleitung, Alternativen oder Kommentare. & \multirow{3}{*}{SFT + DPO} \\
     \hline
     \multirow{4}{*}{3} & Schreibe den folgenden komplexen Satz um und verwende einfachere W\"orter, k\"urzere S\"atze und reduzierte grammatikalische Strukturen. Der Inhalt und die Bedeutung sollen nach dem Umschreiben unver\"andert bleiben. Bitte gib nur die Vereinfachung an, ohne Einleitung, Alternativen oder Kommentare. Komplex: <\texttt{complex\_sentence}>. Leicht: & \multirow{4}{*}{SFT + DPO} \\
     \hline
     \multirow{4}{*}{4} & Formulieren Sie den komplexen Satz um, indem Sie mindestens einen neuen einfachen Satz bilden. Behalten Sie die gleiche Bedeutung des Ausgangssatzes bei. Geben Sie bitte nur die Vereinfachung an, ohne Einleitung, Alternativen oder Kommentare. Komplex: <\texttt{complex\_sentence}>. Leich: & \multirow{4}{*}{SFT + DPO} \\
     \hline 
     \multirow{5}{*}{5} & Schreibe den folgenden komplexen Satz in Leichter Sprache um. Die Vereinfachung soll kurz und von geringer Komplexit\"at sein (durchschnittlich acht bis f\"unfzehn W\"orter pro Satz) und eine geringe Anzahl von Aussagen pro Satz enthalten. Bitte gib nur die Vereinfachung an, ohne Einleitung, Alternativen oder Kommentare. Komplex: <\texttt{complex\_sentence}>. Leicht: & \multirow{5}{*}{SFT + DPO} \\
     \hline 
     \multirow{4}{*}{6} & Schreibe den folgenden komplexen Satz in Leichter Sprache um. Die W\"orter in deiner Vereinfachung sollen kurz, beschreibend, und h\"aufig verwendet von Menschen mit kognitiver Beeintr\"achtigung sein. Bitte gib nur die Vereinfachung an, ohne Einleitung, Alternativen oder Kommentare. Komplex: <\texttt{complex\_sentence}>. Leicht: & \multirow{4}{*}{SFT + DPO} \\
     \hline 
     \multirow{4}{*}{7} & Schreibe den folgenden komplexen Satz in Leichter Sprache um. Deine Vereinfachung soll f\"ur Menschen mit kognitiver Beeintr\"achtigung in \"Osterreich verst\"andlich sein. Bitte gib nur die Vereinfachung an, ohne Einleitung, Alternativen oder Kommentare. Komplex: <\texttt{complex\_sentence}>. Leicht: & \multirow{4}{*}{SFT + DPO} \\
     \hline
     \multirow{7}{*}{8} & Schreiben Sie den folgenden komplexen Satz in Leichter Sprache um. Sie k\"onnen 1) den Satz in mehrere S\"atze aufteilen, 2) Die Wortstellung \"andern, um die Grammatik zu vereinfachen, 3) W\"orter hinzuf\"ugen, um schwierige Konzepte zu erkl\"aren, 4) W\"orter, die sich mit unn\"otigen Informationen zusammenh\"angen, entfernen, und 5) schwierige W\"orter durch einfache Vokabeln ersetzen. Achten Sie darauf, dass der Satz leichter verst\"andlich bleibt, ohne die Bedeutung zu ver\"andern. Bitte geben Sie nur die Vereinfachung an, ohne Einleitung, Alternativen oder Kommentare. Komplex: <\texttt{complex\_sentence}>. Leicht: & \multirow{7}{*}{SFT + DPO} \\
     \hline
     \multirow{4}{*}{9} & Schreibe den folgenden Satz in Leichter Sprache um. Bitte gib nur die Vereinfachung an, ohne Einleitung, Alternativen oder Kommentare. Hier ist ein Beispiel. Komplex: <\texttt{complex\_sentence1}>. Leicht: <simple\_sentence1>. Schreibe deine Vereinfachung nach ``Leicht:''. Komplex: <\texttt{complex\_sentence}>. Leicht: & \multirow{4}{*}{SFT} \\
     \hline 
     \multirow{5}{*}{10} & Schreibe den folgenden komplexen Satz in Leichter Sprache um. Bitte gib nur die Vereinfachung an, ohne Einleitung, Alternativen oder Kommentare. Hier sind zwei Beispiele. Komplex: <\texttt{complex\_sentence1}>. Leicht: <\texttt{simple\_sentence1}>. Komplex: <\texttt{complex\_sentence2}>. Leicht: <\texttt{simple\_sentence2}>. Schreibe deine Vereinfachung nach ``Leicht:''. Komplex: <\texttt{complex\_sentence}>. Leicht: & \multirow{5}{*}{SFT} \\
     \hline
    \end{tabular}
    %}
    \vspace{1mm}
    \caption{\textbf{Prompt templates used for SFT and DPO}. <\texttt{complex\_sentence}> represents the complex text to be simplified, while <\texttt{complex\_sentence1}> and <\texttt{complex\_sentence2}> refer to example complex texts. Correspondingly, <\texttt{simple\_sentence1}> and <\texttt{simple\_sentence2}> serve as their respective simplifications.}
    \label{tab:prompts}
\end{table}

\newpage 

\section{Details of Evaluation of SFT Checkpoints}
\label{sec:training-details}

\begin{figure}[!htb]
    \centering
    \includegraphics[width=0.8\linewidth]{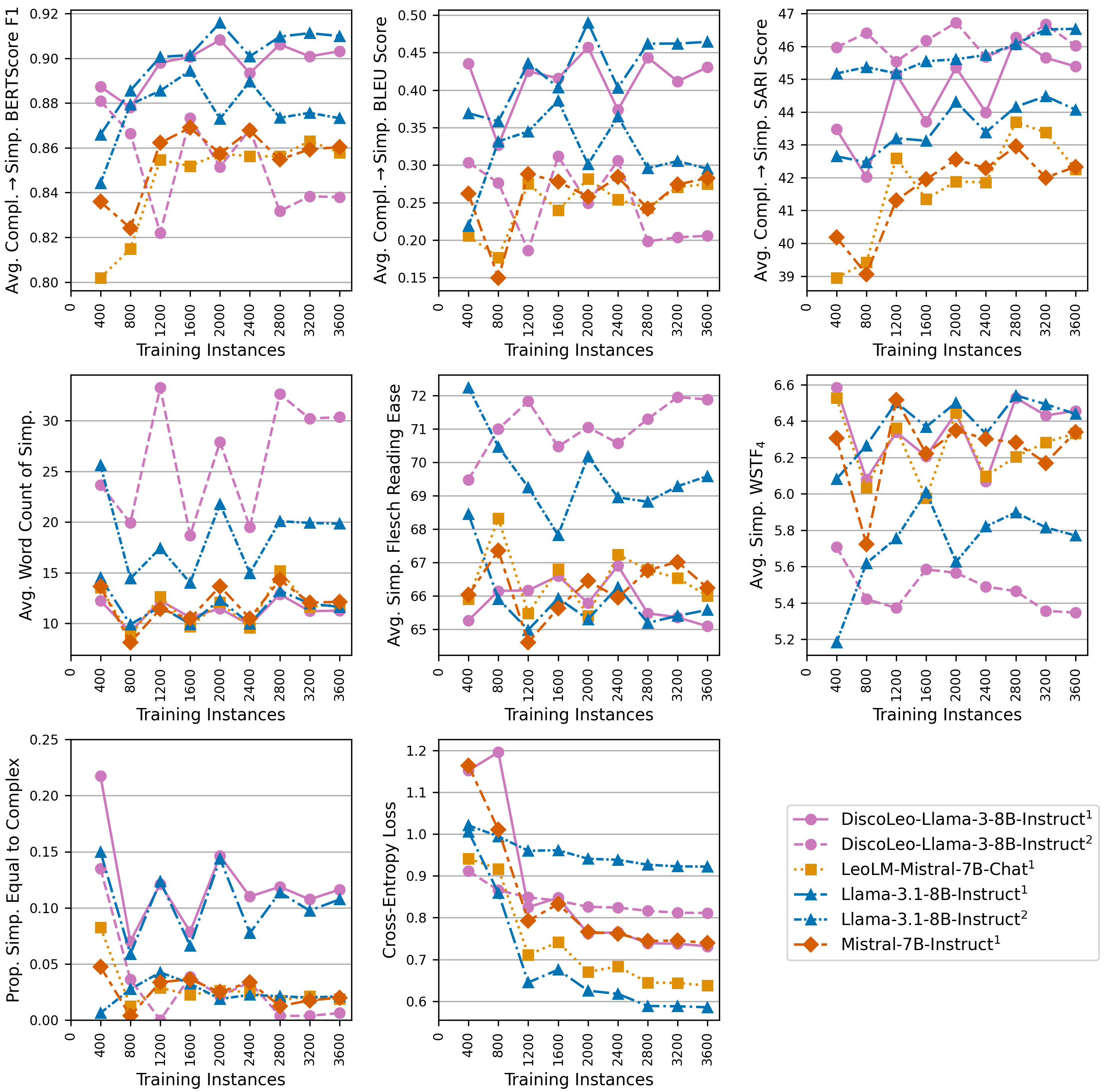}
    \caption{\textbf{Cross-model comparison for SFT checkpoint evaluation.}
    This figure compares different SFT configurations across multiple models using two loss computation strategies. \textbf{Full-prompt loss} (denoted as 1 in the legend) includes both instruction and completion tokens in the loss calculation, while \textbf{completion-only loss} (denoted as 2) considers only the completion tokens. This distinction reflects different assumptions about how supervision signal should be distributed over the input. We selected three different checkpoints for DPO post-training based on this evaluation.}
    \Description{The figure presents a 3×3 grid of line plots comparing nine ATS evaluation metrics for multiple supervised fine-tuning configurations across different models and training sizes. Each subplot has the x-axis labeled “Training Instances” and shows values ranging from 400 to 3200. The y-axis label of each plot corresponds to a different evaluation metric. The top row metrics include: BERTScore F1, BLEU Score, and SARI Score,” each reflecting similarity-based simplification quality. The middle row includes: “Average Word Counts,” “Average Flesch Reading Ease,” and “Average WSTF4,” which capture readability and verbosity. The bottom row includes: “Proportion of Simplifications Equal to Complex," indicating the rate of identity outputs, and “Cross-Entropy Loss,” measuring model loss values. Each plot compares five model configurations differentiated by both color and marker shape. The two models DiscoLeo-Llama-3-8B-Instruct and Llama-3.1-8B-Instruct are each displayed under two variants: one using full-prompt loss, and one using completion-only loss. The other two models displayed are LeoLM-Mistral-7B-Chat and Mistral-7B-Instruct, which are only shown with full-prompt loss. Full-prompt loss variants are drawn with solid or dashed lines and circular or triangular markers, while completion-only loss variants use alternative dashed lines and square or diamond markers. Visual differences in trends are apparent across metrics: for instance, completion-only loss tends to yield lower cross-entropy but may underperform in metrics like BERTScore and SARI. In the lower left plot, all models show a low proportion of outputs equal to the input, with only minor variation. The overall layout enables side-by-side inspection of how supervision signal allocation—either across the full prompt or the completion only—affects simplification quality, readability, and training efficiency.}
    \label{fig:sft_modelmix_grid_search}
\end{figure}

Figure~\ref{fig:sft_modelmix_grid_search} illustrates the performance of SFT checkpoints during development across various metrics. For checkpoint selection, we prioritize models that generate high-quality, readable ATS outputs, as reflected by strong reference-based SARI and reference-free $\text{WSTF}_4$ scores.

To train all SFT models, we employed the AdamW optimizer \citep{loshchilovdecoupled} with a weight decay of 0.01, a cosine annealing learning rate scheduler \citep{loshchilov2022sgdr}, gradient norm clipping at 1, a maximum sequence length of 300 tokens, a batch size of 16, and FP16 mixed precision. We performed parameter-efficient fine-tuning (PEFT; \citep{he2022towards}) on a single NVIDIA A100 GPU using LoRA \citep{hu2022lora} with rank 16, a scaling factor of 32, and a dropout rate of 0.05. Our grid search for hyperparameters scanned across gradient accumulation step size (1, 2, and 4) and learning rate (1e-5, 5e-5, and 1e-4). We selected DiscoLeo-Llama-3-8B-Instruct for hyperparameter optimization due to its abundance of German-language training data relative to Llama-3.1-8B-Instruct and Mistral-7B-Instruct as well as its extensive instruction tuning compared to LeoLM-Mistral-7B-Chat.

\newpage

\section{Details of Preference Pair Creation}

\begin{figure}[htb]
    \centering
    \begin{minipage}{0.68\textwidth}
        \includegraphics[width=\linewidth]{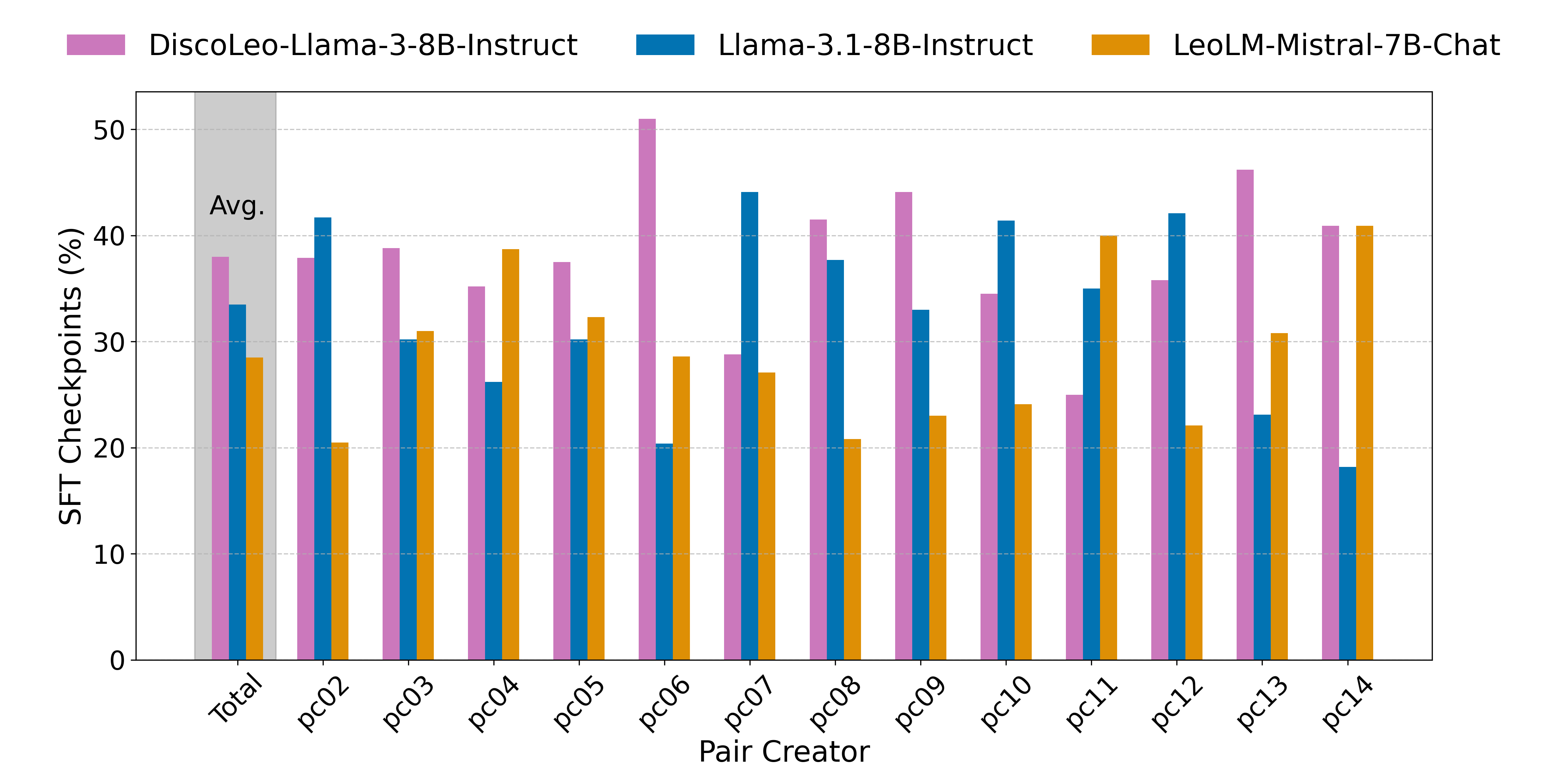}
    \end{minipage}%
    % \hspace{0.3em}
    \begin{minipage}{0.3\textwidth}
    \vspace{-1.3cm}
        \captionof{figure}{\textbf{Distribution of SFT model source for created ATS pairs.} This figure reflects the relative prevalence of different SFT backbone LLMs in the HF4ATS dataset and indicates that human preferences differ from the model perspective. The shaded left bar (Avg.) shows overall averages ranging from approximately 28\% to 37\%, with a plurality of pairs coming from the DiscoLeo-Llama model.}
        \Description{The bar chart compares the distribution of the three SFT model checkpoints—DiscoLeo-Llama-3-8B-Instruct (in magenta color), Llama-3.1-8B-Instruct (in blue color), and LeoLM-Mistral-7B-Chat (in orange color)—that generated the ATS pairs created by each pair creator. The x-axis labels range from pc02 through pc14 and refer to thirteen pair creators. There is also an aggregated "Total" column on the far left. The vertical axis represents the proportion of pairs generated by a given SFT checkpoint, and it ranges between 0 and 50\%. The "Total" column is highlighted with a vertical gray shading and labeled “Avg.” to indicate the overall average generating-model distribution across all created ATS pairs. For about half of pair creators, DiscoLeo-Llama-3-8B-Instruct is the most preferred SFT checkpoint, often responsible for more than 35\% of a creator's pairs. It's prevalence is particularly strong for pc06, pc09, and pc13. Llama-3.1-8B-Instruct prevalence varies more across creators, sometimes ranking highest (e.g., pc02, pc07, pc12), while LeoLM-Mistral-7B-Chat generally trails behind despite a competitive or top performance for pc04, pc11, and pc14.}
        \label{fig:sft-prevalence}
    \end{minipage}
\end{figure}

Figure \ref{fig:sft-prevalence} shows that pair creators exhibited individual preferences for specific SFT checkpoints despite the model-blind pair creation procedure.

\begin{figure}[!htb]
    \centering
    \begin{minipage}{0.35\textwidth}
        \vspace{-0.9cm}
        \captionof{figure}{\textbf{Information-level annotation in the dataset HF4ATS-DPO.} This figure shows the percentage of ATS preference pairs labeled by each pair creator as containing either equal or unequal information, with almost 80\% of all pairs possessing information equality. This categorization reflects whether both options preserve the same content from the original input, offering insight into whether information equity plays an essential role in DPO post-training for ATS.
        }
        \Description{The figure is a horizontal stacked bar chart displaying the proportion of ATS preference pairs labeled as showing either “Inequality” or “Equality” in terms of information preservation. There is one stacked bar for each of the 13 individual pair creators (pc02 to pc14) and one stacked bar indicating the global average. The y-axis lists the pair creators in descending order from pc14 at the top to pc02 near the bottom, with the additional “Total” row at the very bottom. The x-axis represents information level percentages from 0 to 100. Each row contains two color-coded bars: a dark gray segment representing the percentage of preference pairs labeled as “Inequality” and a green segment representing “Equality.” For instance, pc12 has the highest inequality proportion at 65.2\%, while pc13 has 0\% inequality and 100\% equality. The total row at the bottom summarizes the overall distribution across all pair creators, with 22.4\% of examples marked as information-unequal and 77.6\% as information-equal. Each bar segment is annotated with the corresponding percentage. The background of the total row is shaded to visually separate it from individual contributors. A legend at the top identifies the color scheme, where gray denotes “Inequality” and green denotes “Equality.” This visualization highlights the variability in how pair creators perceive semantic equivalence across simplification candidates and provides insight into how information equity may impact DPO post-training outcomes.}
        \label{fig:info_level}
    \end{minipage}%
    \hspace{1em}
    \begin{minipage}{0.55\textwidth}
        \includegraphics[width=\linewidth]{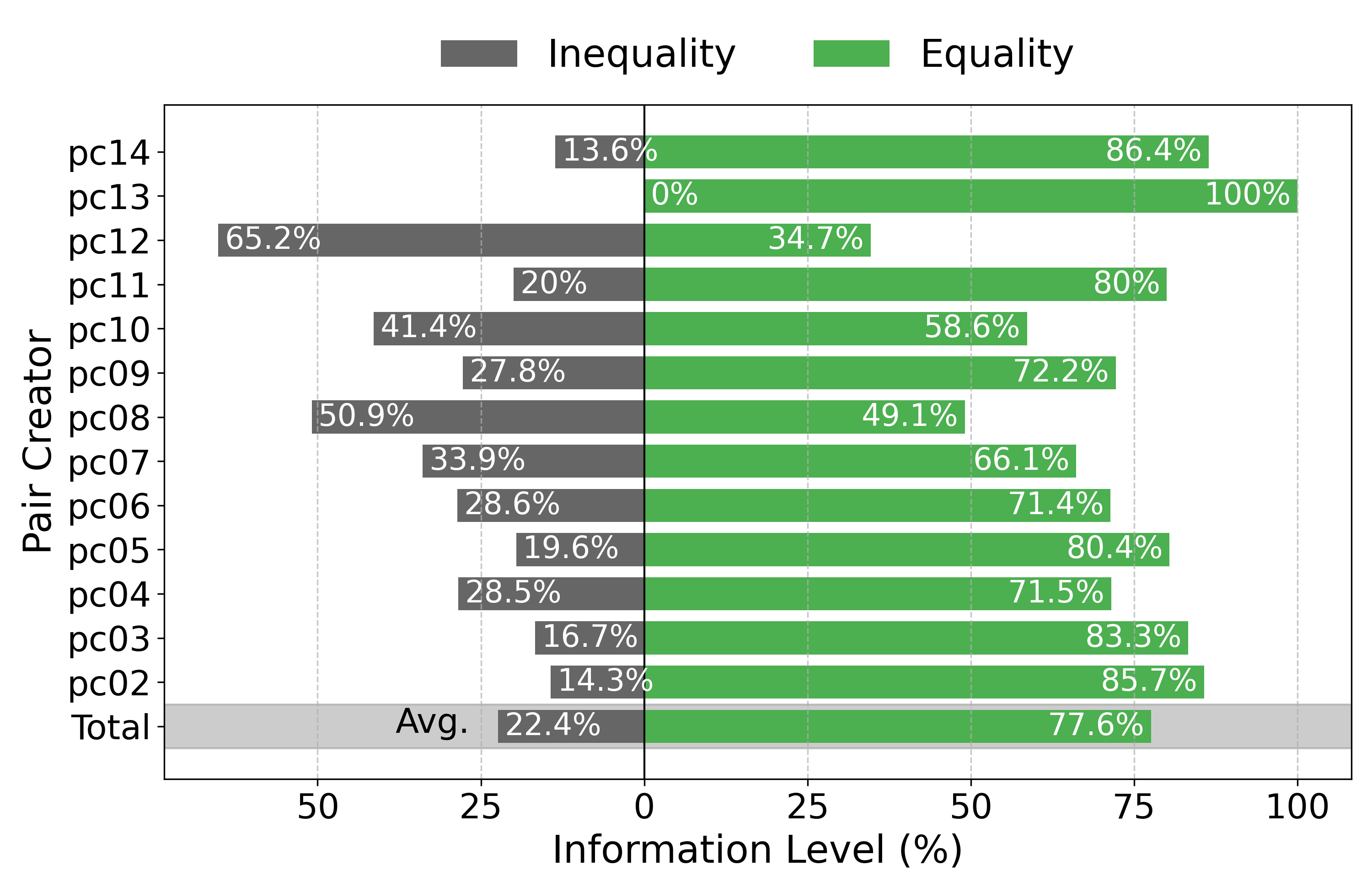}
    \end{minipage}
\end{figure}

Figure~\ref{fig:info_level} presents the percentage of pairs labeled as having equal or differing information, grouped by pair creator. The shaded bottom row (Avg.) shows the overall average, with nearly 80\% of pairs labeled as having equal information.

\begin{figure}[H]
    \centering
    \begin{minipage}{0.68\textwidth}
        \includegraphics[width=\linewidth]{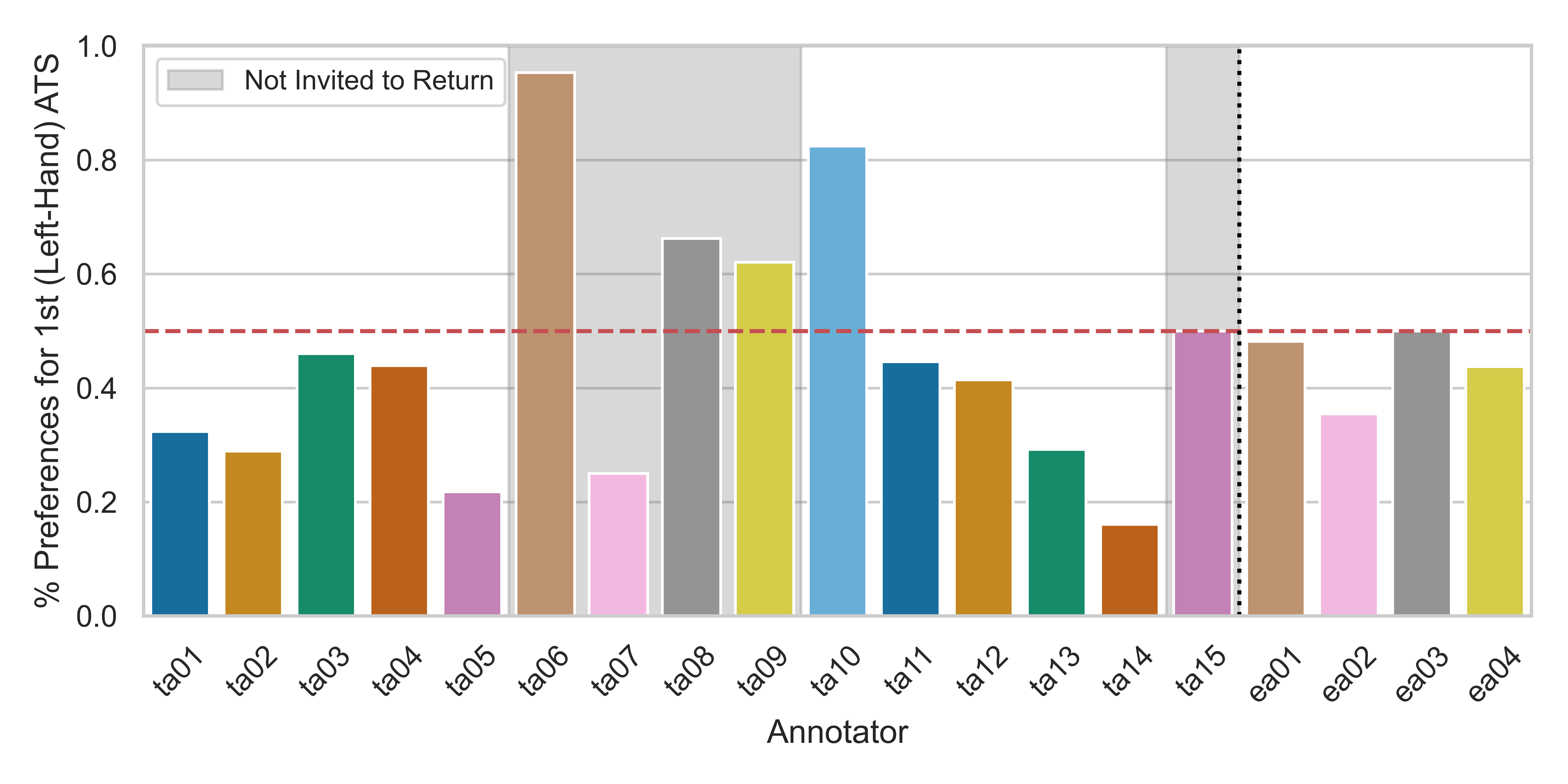}
    \end{minipage}%
    % \hspace{0.3em}
    \begin{minipage}{0.3\textwidth}
    \vspace{-1.3cm}
        \captionof{figure}{Preference rate for the left-hand option by user. Annotators such as ta06 exhibited an overwhelming preference for one side, suggesting they did not understand or adhere to task instructions. Apart from concentrating preferences on one side, annotators may not have been invited to return for other reasons (e.g. admitting they struggled to understand the task).}
        \Description{}
        \label{fig:anno-side}
    \end{minipage}
\end{figure}

Figure~\ref{fig:anno-side} suggests that some target group annotators consistently favored one side, indicating possible non-adherence to task instructions.

\end{document}